%% file: emnlp-ijcnlp-2019.tex
\documentclass[11pt,a4paper]{article}
\usepackage[hyperref]{emnlp-ijcnlp-2019}
\usepackage{times}
\usepackage{latexsym}

\usepackage{url}
\usepackage{graphicx}
\usepackage{caption}
\usepackage[utf8x]{inputenc}
\usepackage{amsmath}
\usepackage{tabularx}
\usepackage{mathtools}
\usepackage{booktabs}
\usepackage{url}
\usepackage{longtable}
\usepackage{tabu}
\usepackage{algorithm}
\usepackage{subfigure}
\usepackage{bbm}
\usepackage{xcolor}
\usepackage[noend]{algpseudocode}
\usepackage[normalem]{ulem}
\makeatletter
\def\BState{\State\hskip-\ALG@thistlm}
\makeatother

\DeclareMathOperator*{\argmin}{arg\,min}

\aclfinalcopy % Uncomment this line for the final submission
\title{A Little Annotation does a Lot of Good: \\A Study in Bootstrapping Low-resource Named Entity Recognizers}

\author{Aditi Chaudhary, \hspace{2mm}
    Jiateng Xie, \hspace{2mm}
  Zaid Sheikh,  \\
   {\bf Graham Neubig,}  \hspace{2mm}
  {\bf Jaime G. Carbonell} \\
  \texttt{$\{aschaudh, jiatengx, zsheikh, gneubig, jgc\}$@cs.cmu.edu} \\
  Language Technologies Institute \\
  Carnegie Mellon University
 }

\begin{document}
\maketitle
\begin{abstract}
  Most state-of-the-art models for named entity recognition (NER) rely on the availability of large amounts of labeled data, making them challenging to extend to new, lower-resourced languages. However, there are now several proposed approaches involving either cross-lingual transfer learning, which learns from other highly resourced languages, or active learning, which efficiently selects effective training data based on model predictions. This paper poses the question: given this recent progress, and limited human annotation, what is the most effective method for efficiently creating high-quality entity recognizers in under-resourced languages? Based on extensive experimentation using both simulated and real human annotation, we find a dual-strategy approach best, starting with a cross-lingual transferred model, then performing targeted annotation of only uncertain entity spans in the target language, minimizing annotator effort.
Results demonstrate that cross-lingual transfer is a powerful tool when very little data can be annotated, but an entity-targeted annotation strategy can achieve competitive accuracy quickly, with just one-tenth of training data. The code is publicly available here.\footnote{\url{https://github.com/Aditi138/EntityTargetedActiveLearning}}
\end{abstract}
\input{intro.tex}

\input{method.tex}
\input{experiments.tex}

\input{relatedwork.tex}
\input{conclusion.tex}

\section*{Acknowledgement}
The authors would like to thank Sachin Kumar, Kundan Krishna, Aldrian Obaja Muis, Shirley Anugrah Hayati, Rodolfo Vega and Ramon Sanabria for participating in the human annotation experiments.
This work is sponsored by Defense Advanced Research
Projects Agency Information Innovation Office
(I2O). Program: Low Resource Languages
for Emergent Incidents (LORELEI). Issued by
DARPA/I2O under Contract No. HR0011-15-C0114.
The views and conclusions contained in this
document are those of the authors and should not
be interpreted as representing the official policies,
either expressed or implied, of the U.S. Government.
The U.S. Government is authorized to reproduce
and distribute reprints for Government purposes
notwithstanding any copyright notation here
on.

\bibliography{emnlp-ijcnlp-2019}
\bibliographystyle{acl_natbib}

\newpage
\input{appendix.tex}

\end{document}

%% file: intro.tex
\section{Introduction}
\label{intro}
Named entity recognition (NER) is the task of detecting and classifying named entities in text into a fixed set of pre-defined categories (person, location, etc.) with several downstream applications including machine reading \cite{chen2017reading}, entity and event co-reference \cite{yang2016joint}, and text mining \cite{han2012entity}. Recent advances in deep learning have yielded state-of-the-art performance on many sequence labeling tasks, including NER \cite{collobert2011natural,state,lample2016neural,peters2018deep}. However, the performance of these models is highly dependent on the availability of large amounts of annotated data, and as a result their accuracy is significantly lower on languages that have fewer resources than English. In this work, we ask the question ``how can we efficiently bootstrap a high-quality named entity recognizer for a low-resource language with only a small amount of human effort?''
\begin{figure*}
\small
\includegraphics[width=\textwidth]{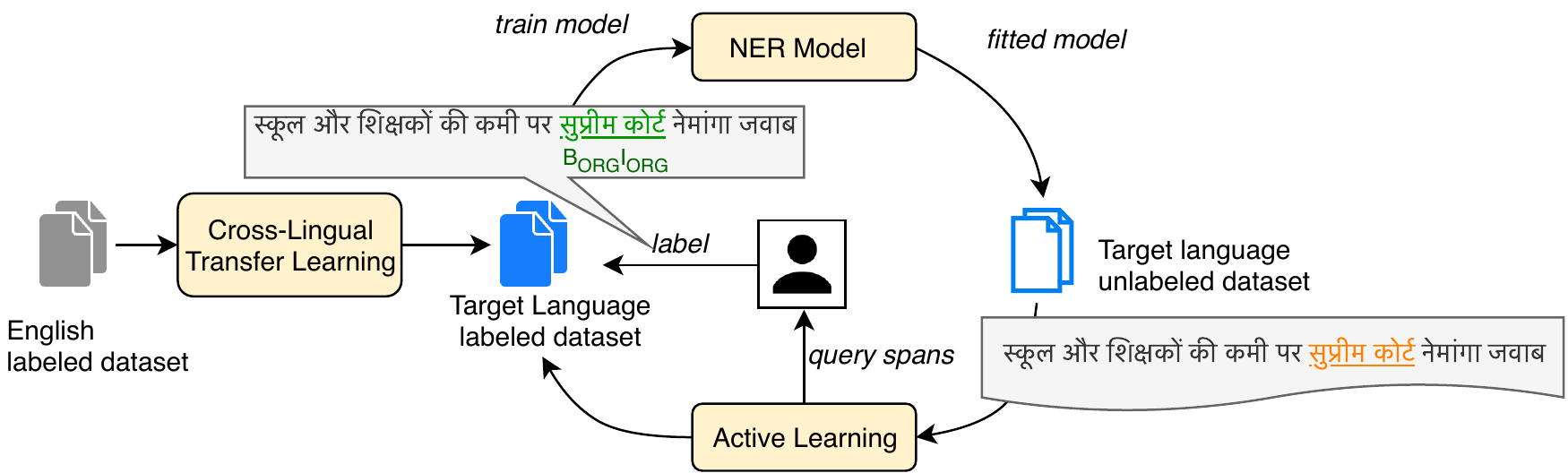}
\caption{\label{system} Our proposed recipe: cross-lingual transfer is used for projecting annotations from an English labeled dataset to the target language. Entity-targeted active learning is then used to select informative sub-spans which are likely entities for humans to annotate. Finally, the NER model is fine-tuned on this partially-labeled dataset. }
\end{figure*}
Specifically, we leverage recent advances in data-efficient learning for low-resource languages, proposing the following ``recipe'' for bootstrapping low-resource entity recognizers:
First, we use \emph{cross-lingual transfer learning} \cite{yarowsky2001inducing, ammar2016massively}, which applies a model trained on another language to low-resource languages, to provide a good preliminary model to start the bootstrapping process.
Specifically, we use the model of \newcite{xie2018neural}, which reports strong results on a number of language pairs.
Next, on top of this transferred model we further employ \emph{active learning} \cite{settles2008analysis, marcheggiani2014experimental}, which helps improve annotation efficiency by using model predictions to select informative, rather than random, data for human annotators.
Finally, the model is \emph{fine-tuned} on data obtained using active learning to improve accuracy in the target language.

Within this recipe, the choice of specific method for choosing and annotating data within active learning is highly important to minimize human effort.
One relatively standard method used in previous work on NER is to select full sequences based on a criterion for the uncertainty of the entities recognized therein \cite{culotta2005reducing}.
However, as it is often the case that only a single entity within the sentence may be of interest, it can still be tedious and wasteful to annotate full sequences when only a small portion of the sentence is of interest~\citep{neubig2011pointwise, sperber2014segmentation}.
Inspired by this finding and considering the fact that named entities are both important and sparse, we propose an entity-targeted strategy to save annotator effort. Specifically, we select uncertain subspans of tokens within a sequence that are most likely named entities. This way, the annotators only need to assign types to the chosen subspans without having to read and annotate the full sequence. To cope with the resulting partial annotation of sequences, we apply a constrained version of conditional random fields (CRFs), partial CRFs, during training that only learn from the annotated subspans ~\citep{tsuboi2008training,wanvarie2011active}.

To evaluate our proposed methods, we conducted simulated active learning experiments on 5 languages: Spanish, Dutch, German, Hindi and Indonesian.
%from the CoNLL 2002 and 2003 datasets and Hindi, Indonesian and  Spanish from the LDC language packs. %We use the CoNLL 2003 English training data to perform cross-lingual transfer for all our experiments. 
Additionally, to study our method in a more practical setting, we conduct human annotation experiments on two low-resource languages, Indonesian and Hindi, and one simulated low-resource language, Spanish. %, from LDC. 
  In sum, this paper makes the following contributions:
\begin{enumerate}
  \item We present a bootstrapping recipe for improving low-resource NER. With just one-tenth of tokens annotated, our proposed entity-targeted active learning method provides the best results among all active learning baselines, with an average improvement of 9.9 F1.
  
    \item Through simulated experiments, we show that cross-lingual transfer is a powerful tool, outperforming the un-transferred systems by an average of 8.6 F1 with only one-tenth of tokens annotated.
  
    \item Human annotation experiments show that annotators are more accurate in annotating entities when using the entity-targeted strategy as opposed to full sequence annotation. Moreover, this strategy minimizes annotator effort by requiring them to label fewer tokens than the full-sequence annotation.
\end{enumerate}

%% file: method.tex
\section{Approach}
\label{method}
As noted in the introduction, our bootstrapping recipe consists of three components (1) cross-lingual transfer learning, (2) active learning to select relevant parts of the data to annotate, and (3) fine-tuning of the model on these annotated segments.
Steps (2) and (3) are continued until the model has achieved an acceptable level of accuracy, or until we have exhausted our annotation budget. The system overview can be seen in Figure \ref{system}.
In the following sections, we describe each of these three steps in detail.

\subsection{Cross-lingual Transfer Learning}
\label{ct}
The goal of cross-lingual learning is to take a recognizer trained in a source language, and transfer it to a target language.
Our approach to doing so for NER follows that of ~\citet{xie2018neural}, and we provide a brief review in this section.

To begin with, we assume access to two sets of pre-trained monolingual word embeddings in the source and target languages, $X$ and $Y$, one small bilingual lexicon, either provided or obtained in an unsupervised manner ~\citep{artetxe2017learning,lample2018word}, and labeled training data in the source language. Using these resources, we train bilingual word embeddings (BWE) to create a word-to-word translation dictionary, and finally use this dictionary to translate the source training data into the target language, which we use to train an NER model.

To learn BWE, we first obtain a linear mapping $W$ by solving the following objective:
\[W^* = \argmin_W \|WX_D - Y_D\|_F \text{~~s.t.~~} W W^\top = I,\]
where $X_D$ and $Y_D$ correspond to the aligned word embeddings from the bilingual lexicon. $F$ denotes the Frobenius norm. We can first compute the singular value decomposition $Y_D^T X_D = U\sum V^\top$, and solve the objective by taking $W^* = U V^\top$. We obtain BWE by linearly transforming the source and target monolingual word embeddings with $U$ and $V$, namely $XU$ and $YV$.

After obtaining the BWE, we find the nearest neighbor target word for every source word in the BWE space using the cross-domain similarity local scaling (CSLS) metric ~\citep{conneau2017word}, which produces a word-to-word translation dictionary. We use this dictionary to translate the source training data into the target language, and simply copy the label for each word, which yields transferred training data in the target language. We train an NER model on this transferred data as our preliminary model. Going forward, we refer to the use of cross-lingual transferred data as \textsc{\small{CT}}.

\subsection{Entity-Targeted Active Learning}
\label{activeLearning}
After training a model using cross-lingual transfer learning, we start the active learning process based on this model's outputs. We begin by training a NER model $\Theta$ using the above model's outputs as training data.  Using this trained model, our proposed entity-targeted active learning strategy, referred as \textsc{\small{ETAL}}, then selects the most informative spans from a corpus $D$ of unlabeled sequences. Given an unlabeled sequence $s$, \textsc{\small{ETAL}} first selects a span of tokens $s_i^j = s_i \cdots s_j $ such that $s_i^j$ is a likely named entity, where $i,j \in [0,|s|]$. Then, in order to obtain highly informative spans across $D$, \textsc{\small{ETAL}} computes the entropy $H$ for each occurrence of the span $s_i^j$ in $D$ and then aggregates them over the entire corpus $D$, given by:
\[ H_{aggregate}(s_i^j) =\sum_{x_i^j\in D} H(x_i^j) \mathbbm{1}(x_i^j = s_i^j) \] where $x$ is an unlabeled sequence in $D$. Finally, the spans $s_i^j$ with the highest aggregate uncertainty $H_{aggregate}$ are selected for manual annotation. 

We now describe the procedure for calculating $H(x_i^j)$, which is the entropy of a span $x_i^j$ being a likely entity.  Given an unlabeled sequence $x$, the trained NER model $\Theta$ is used for computing the marginal probabilities $p_\theta(y_i|\mathbf{x})$ for each token $x_i$ across all possible labels $y_i \in Y$ using the forward-backward algorithm \cite{rabiner1989tutorial}, where $Y$ is the set of all labels. Using these marginals we calculate the entropy of a given span $x_i^j$ being an entity as shown in Algorithm \ref{euclid}.

\begin{algorithm}
\caption{Entity-Targeted Active Learning}\label{euclid}
\begin{algorithmic}[1]
%\State $\textit{s} \gets \textit{unlabeled sentence }$
\State $\textit{B} \gets \textit{label-set denoting beginning of an entity}$
\State $\textit{I} \gets \textit{label-set denoting inside of an entity}$
\State $\textit{O} \gets \textit{outside of an entity span}$
\State $p_\theta(y_i |\mathbf{x}) \gets \textit{marginal probability of label } y_i$ 
\State \textit{ for token } $x_i$\\

\For {$i \gets  1... len(x), j=1 $}
\State $p^{ij}_{span} = \sum_{y \in B}p_\theta(y_i |\mathbf{x})$ 
\For {$j \gets  i+1... len(x) $}
\State $p_{entity} = p^{ij}_{span} * p_\theta(O_j |\mathbf{x})  $
\State $H=\textit{entropy (}p_{entity}\textit{)}$
\If {$H > H_{threshold}$} 
\State $H_{aggregate}(x_i^j) += H$
\EndIf
\State $p^{ij}_{span} = p^{ij}_{span} * \sum_{y \in I} p_\theta(y_j |\mathbf{x})$
\EndFor 
\EndFor 
\end{algorithmic}
\end{algorithm}

Let $B$ denote the set of labels indicating beginning of an entity,  $I$ the set of labels indicating inside of an entity and $O$ denoting outside of an entity. First, we compute the probability of a span $x_i^j$  being an entity, starting with the token $i$, by marginalizing $p_\theta(y_i|\mathbf{x})$ over all labels in $B$, denoted as $p^{ij}_{span}$. Since an entity can span multiple tokens, for each subsequent token $j$ being part of that entity, we marginalize $p_\theta(y_j|\mathbf{x})$ over all labels in $I$ and combine it with $p^{ij}_{span}$. Finally, we compute $p_{entity} = p^{ij}_{span} * p_\theta(O_j|\mathbf{x})$, which denotes end of a likely entity. Since we use the marginal probability for computing $p_{entity}$, it already factors in the transition probability between tags. Thus, any invalid sequences such as \textsc{\small{$B_{PER} I_{ORG}$}} have low scores. Since contiguous spans have overlapping tokens, using dynamic programming (DP) to compute $p^{ij}_{span}$ avoids an exponential computation when considering all possible spans in a sequence.  Using $p_{entity}$, we compute the entropy $H$ and only consider the spans having $H$ higher than a pre-defined threshold $H_{threshold}$. The reason for this thresholding is purely for computational purposes as it allows us to discard all spans that have a very low probability of being an entity, keeping the number of spans actually stored in memory low.  As mentioned above, we aggregate the entropy of spans $H_{aggregate}$ over the entire unlabeled set, thus combining uncertainty sampling with a bias towards high frequency entities. 

Using this strategy, we select subspans in each sequence for annotation. The annotator only needs to assign named entity types to the chosen subspans, adjust the span boundary if needed, and ignore the rest of the sequence, saving much effort. 

\subsection{Training the NER model}
\label{model}

With the newly obtained training data from active learning, we attempt to improve the original transferred model. In this section, we first describe our model architecture, and try to address: 1) how to train the NER model effectively with partially annotated sequences? 2) what training scheme is best suited to improve the transferred model?

\subsubsection{Model Architecture}

Our NER model is a BiLSTM-CNN-CRF model based on \citet{state} consisting of: a character-level CNN, that allows the model to capture subword information; a word-level BiLSTM, that consumes word embeddings and produces context sensitive hidden representations; and a linear-chain CRF layer that models the dependency between labels for inference. We use the above model for training the initial NER model on the transferred data as well as for re-training the model on the data acquired from active learning.

\subsubsection{\textsc{Partial-CRF}}
\label{pcrf}
Active learning with span-based strategies such as \textsc{\small{ETAL}}, produces a training dataset of partially labeled sequences. To train the NER model on these partially labeled sequences, we take inspiration from \citet{bellare2007learning, tsuboi2008training} and use a constrained CRF decoder. Normally, CRF computes the likelihood of a label sequence $\mathbf{y}$ given a sequence $\mathbf{x}$ as follows:
\[    p_\theta(\mathbf{y}|\mathbf{x}) = \frac{\prod_{t=1}^{T} \psi_i(y_{t-1},y_t,\mathbf{x},t)}{Z(\mathbf{x})} \]
\[    Z(\mathbf{x}) = \sum_{\mathbf{y} \in \mathbf{Y}(T)} \prod_{t=1}^{T} \psi_i(y_{t-1},y_t,\mathbf{x},t)\]

where $T$ is the length of the sequence, $\mathbf{Y}(T)$ denotes the set of all possible label sequences with length $T$, and $\psi_i(y_{t-1},y_t, \mathbf{x})=\exp(\mathbf{W}^{T}_{y_{t-1},y_t}\mathbf{x}_i + \mathbf{b}_{y_{t-1},y_t})$ is the energy function. To compute the likelihood of a sequence where some labels are unknown, we use a constrained CRF which marginalizes out the un-annotated tokens. Specifically, let $\mathbf{Y_L}$ denote the set of all possible sequences that include the partial annotations (for un-annotated tokens, all labels are possible), and 
we compute the likelihood as:
$p_\theta(\mathbf{Y_L}|\mathbf{x}) = \sum_{\mathbf{y} \in \mathbf{Y_L}} p_\theta(\mathbf{y} | \mathbf{x})$,
referred as \textsc{\small{Partial-CRF}}. 
 
\subsubsection{Training Scheme}
\label{ts}
 To improve our model with the newly labeled data, we directly fine-tune the initial model, trained on the transferred data, on the data acquired through active learning, referred as \textsc{\small{FineTune}}. Each active learning run produces more labeled data, for which this training procedure is repeated again.   We also compare the NER performance using two other training schemes: \textsc{\small{CorpusAug}}, where we train the model on the concatenated corpus of transferred data and the newly acquired data, and \textsc{\small{CorpusAug+FineTune}}, where we additionally fine-tune the model trained using \textsc{\small{CorpusAug}} on just the newly acquired data.

%% file: experiments.tex
\section{Experiments}
\label{exp}
In this section, we evaluate the effectiveness of our proposed strategy in both simulated (\S \ref{simulation}) and human-annotation experiments (\S \ref{human}).% We first describe the experimental settings.

\subsection{Experimental Settings}
\label{expsetting}
\paragraph{Datasets:} The first evaluation set includes the benchmark CoNLL 2002 and 2003 NER datasets \cite{TjongKimSang, sang2003introduction} for Spanish (from the Romance family), Dutch and German (like English, from the Germanic family).  We use the standard corpus splits for train/dev/test. The second evaluation set is for the low-resource setting where we use the Indonesian (from the Austronesian family), Hindi (from the Indo-Aryan family) and Spanish datasets released by the Linguistic Data Consortium (LDC).\footnote{LDC2017E62,LDC2016E97,LDC2017E66} We generate the train/dev/test split by random sampling.  Details of the corpus statistics are in the Appendix \S \ref{corpustable}.

\paragraph{English-transferred Data:} We use the same experimental settings and resources as described in \citet{xie2018neural} to  get the translations of the English training data for each target language.

 \paragraph{Active Learning Setup:} As described in Section \S \ref{activeLearning}, a DP-based algorithm is employed to select the uncertain entity spans which runs for all n-grams having length $<=$ 5. This length was approximated by computing the 90th percentile on the length of  entities in the English training data. $H_{threshold}$ is a hyper-parameter  set to 1e-8. The details of the NER model hyper-parameters can be found in the Appendix \S \ref{nermodel} .

\subsection{Simulation Experiments}
\label{simulation}
\paragraph{Setup:} We use cross-lingual transfer (\S\ref{ct}) to train our initial NER model and test on the target language.  This is the same setting as \citet{xie2018neural} and serves as our baseline.
Then we use several active learning strategies to select data for manual annotation using this trained NER model. We compare our proposed \textsc{\small{ETAL}} strategy  with the following baseline strategies:
\paragraph{\textsc{\small{SAL}}:}  Select whole sequences for which the model has least confidence in the most likely labeling \cite{culotta2005reducing}. 
\paragraph{\textsc{\small{CFEAL}}:}  Select least confident spans within a sequence using the confidence field estimation method \cite{culotta2004confidence}.
\paragraph{\textsc{\small{RAND}}:} Select spans randomly from the unlabeled set for annotation. 

In this experimental setting, we simulate manual annotation by using gold labels for the data selected by active learning. At each subsequent run, we annotate 200 tokens and fine-tune the NER model on all the data acquired so far, which is then used to select data for the next run of annotation.

\subsubsection{Results}
\begin{figure*}%
\centering
\subfigure{%
\label{ex3-dc}%
\includegraphics[width=0.33\textwidth]{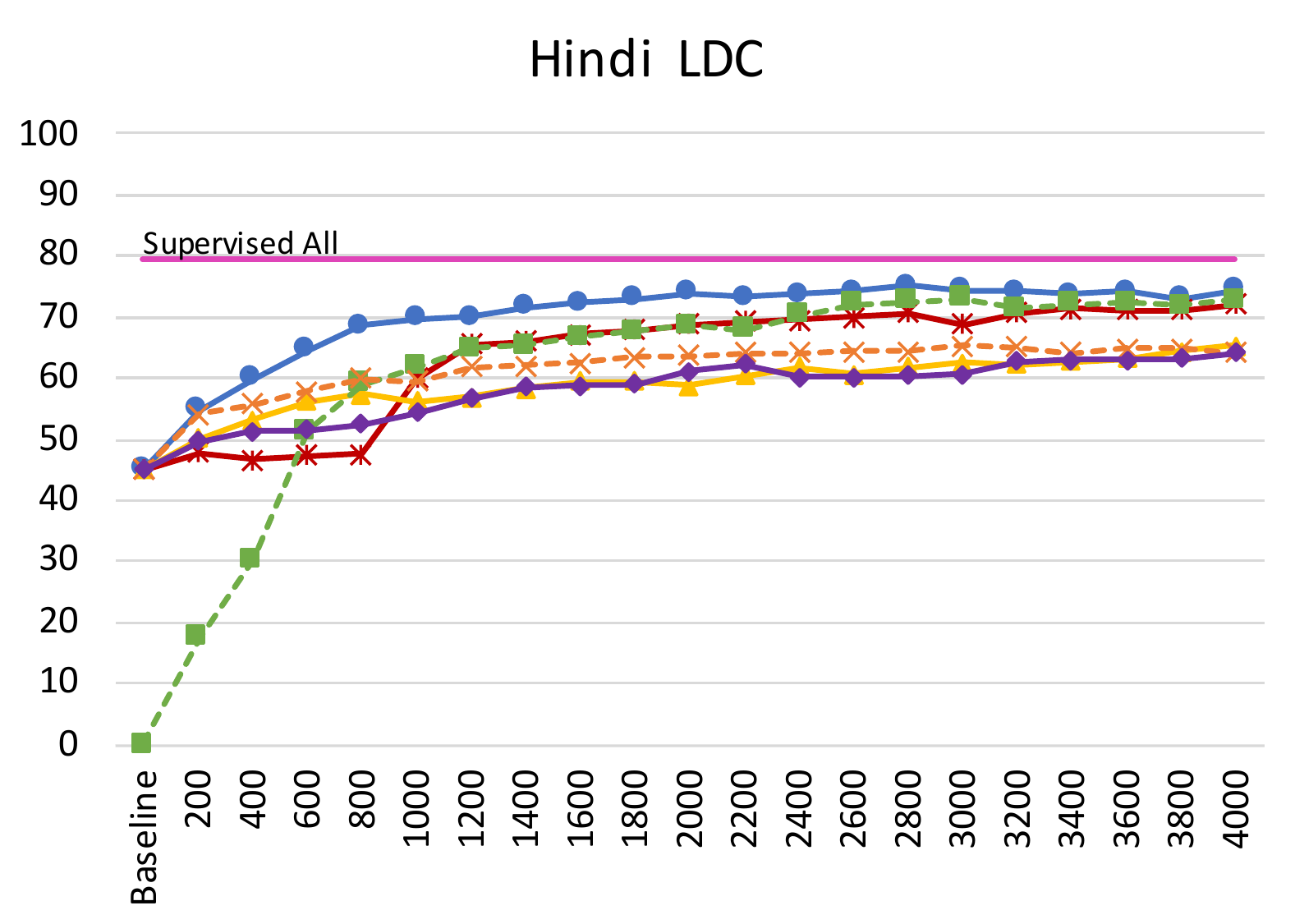}}%\\
~
%\hspace{8pt}%
\subfigure{%
\label{ex3-ed}%
\includegraphics[width=0.33\textwidth]{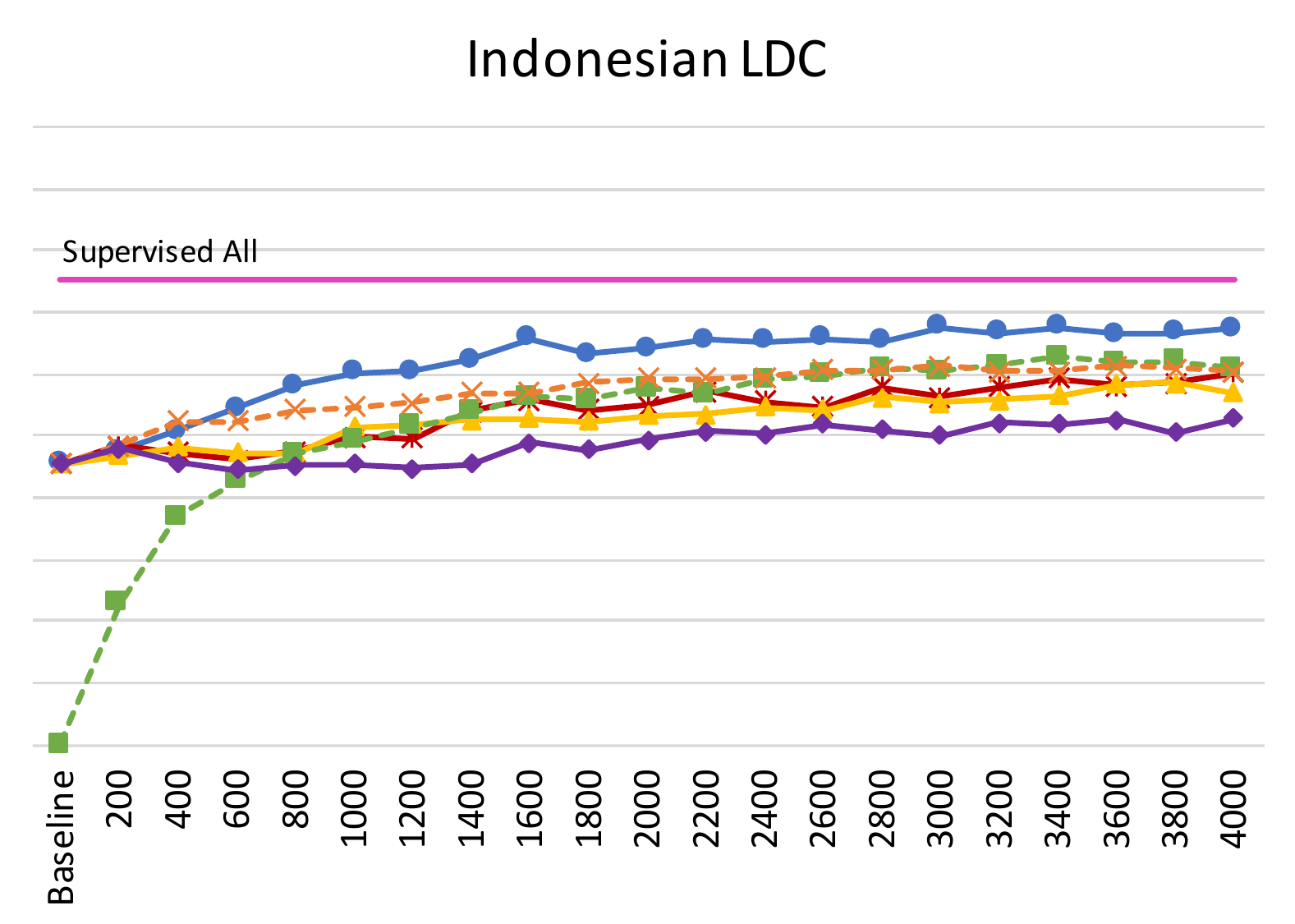}}%\\
~
\subfigure{%
\label{ex3-fe}%
\includegraphics[width=0.33\textwidth]{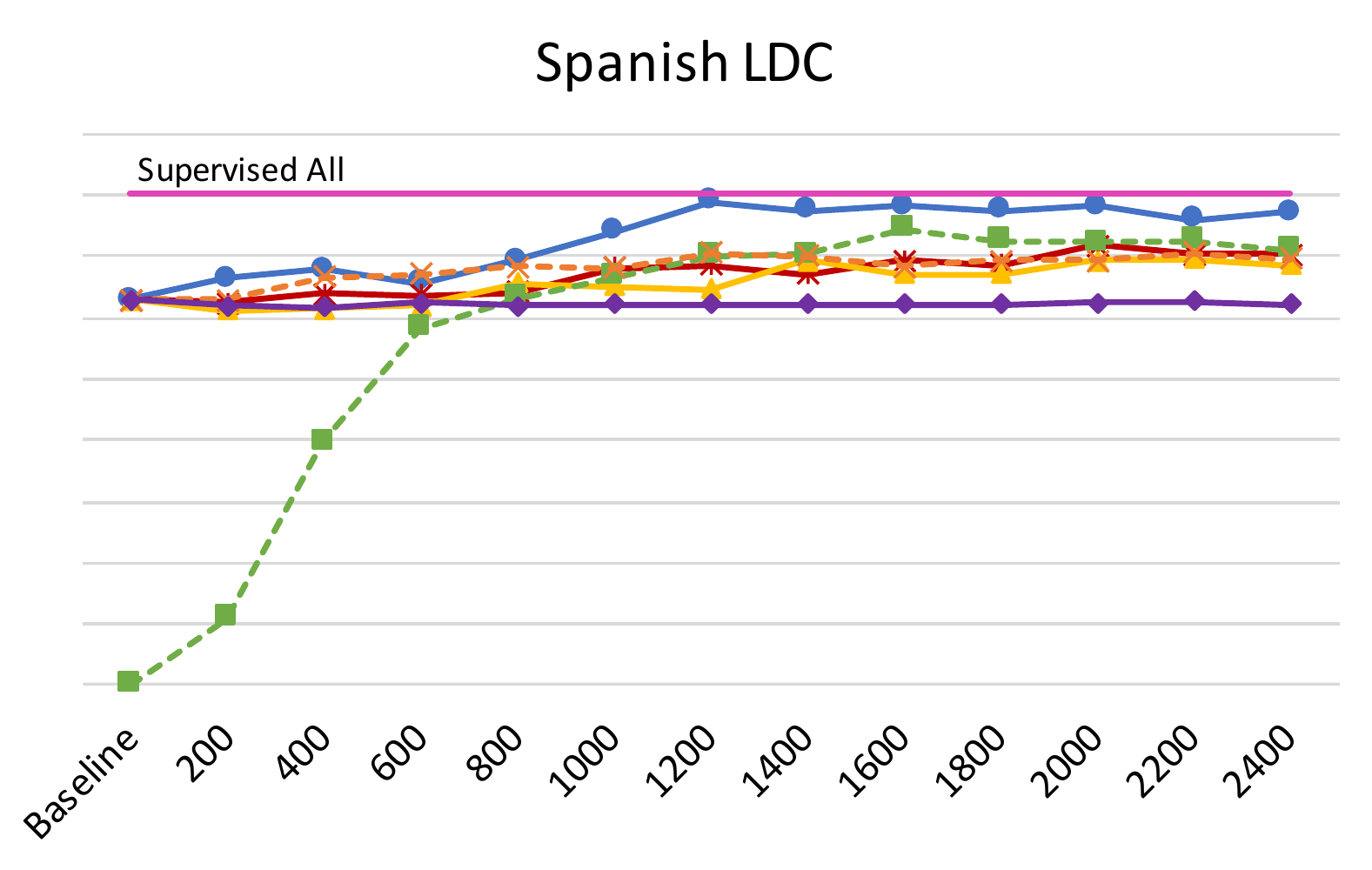}}
%~
\subfigure{%
\label{ex3-gf}%
\includegraphics[width=0.33\textwidth]{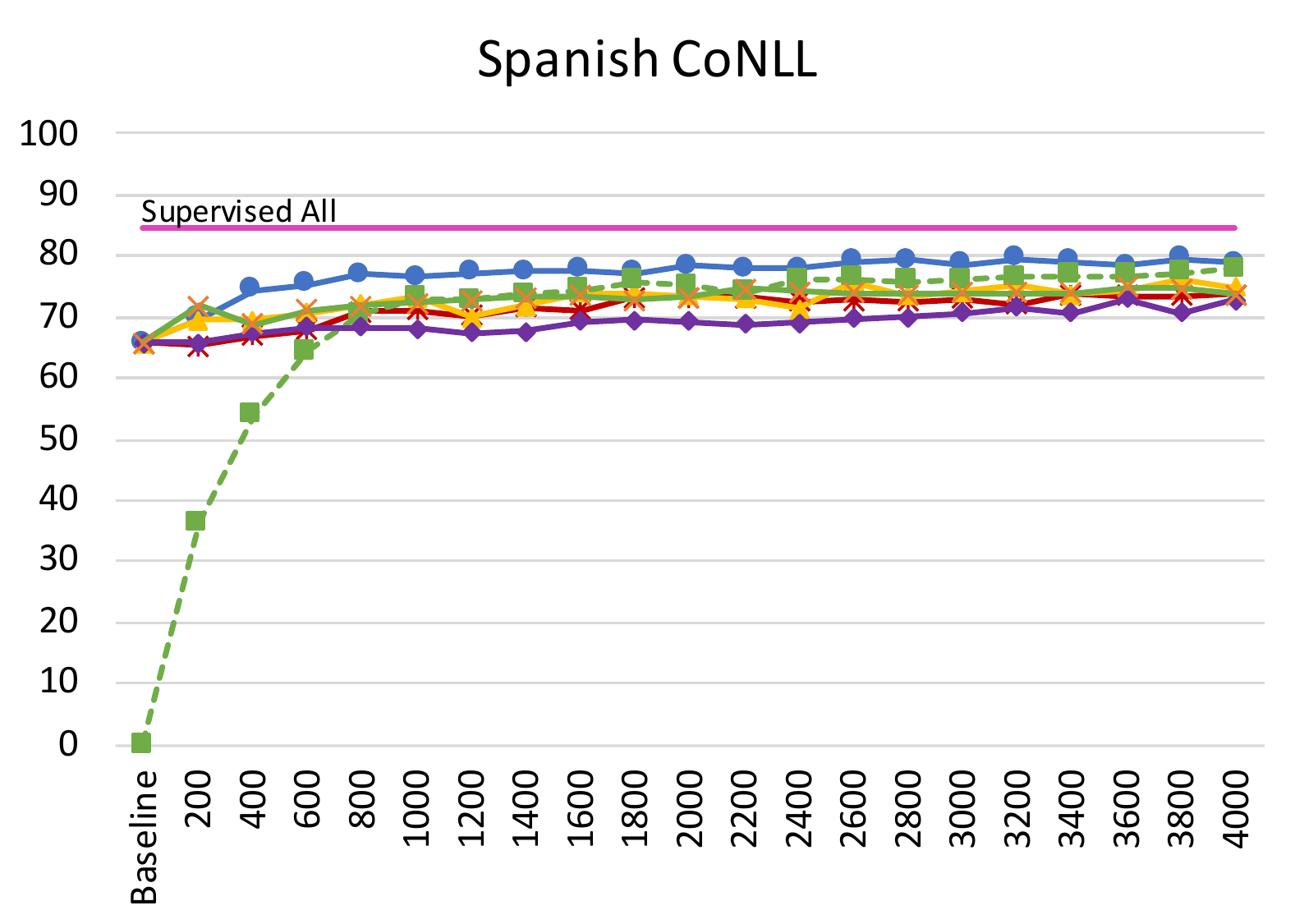}}%\\
~
\subfigure{%
\label{ex3-ha}%
\includegraphics[width=0.33\textwidth]{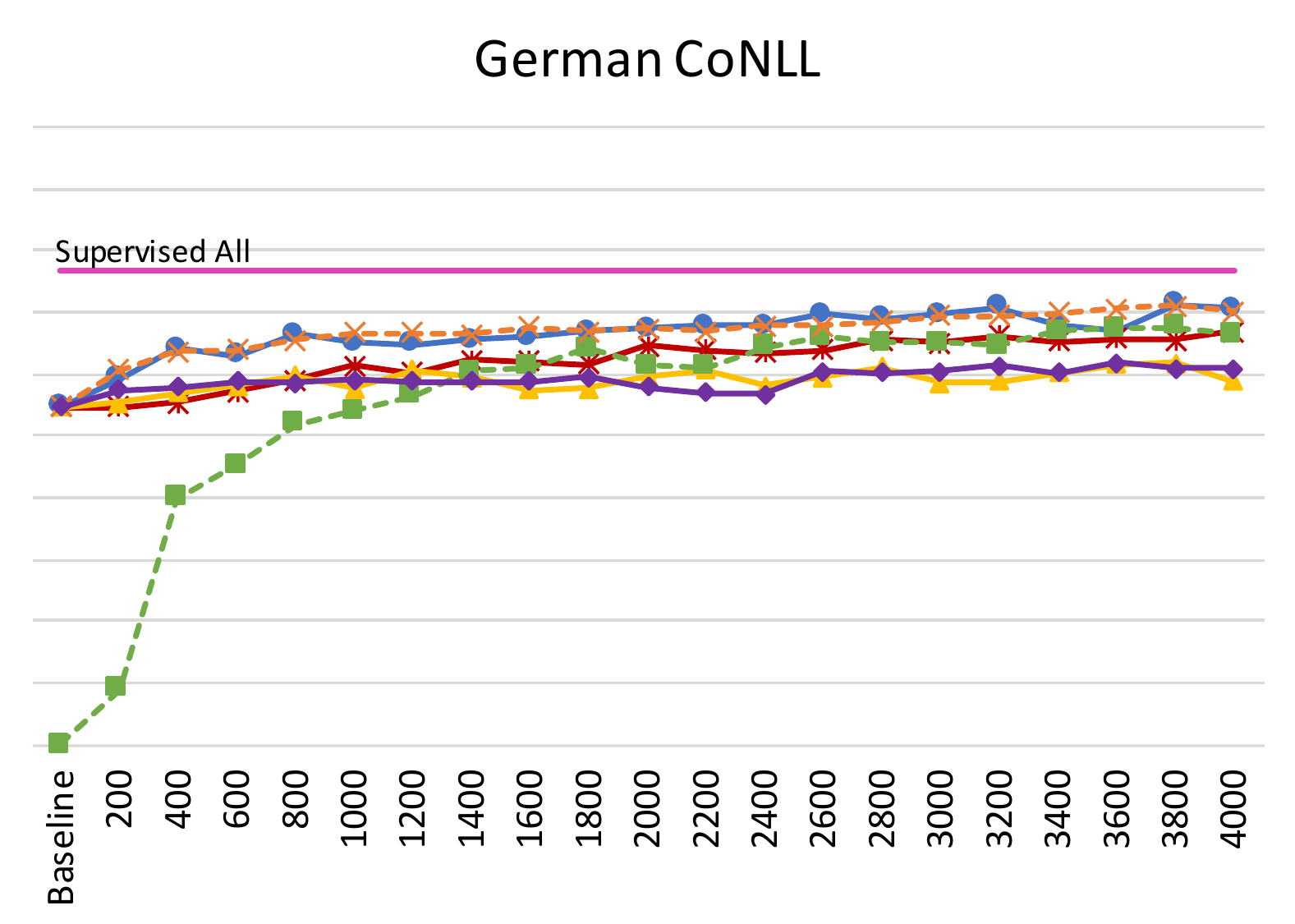}}%\\
~
%\hspace{8pt}%
\subfigure{%
\label{ex3-ib}%
\includegraphics[width=0.33\textwidth]{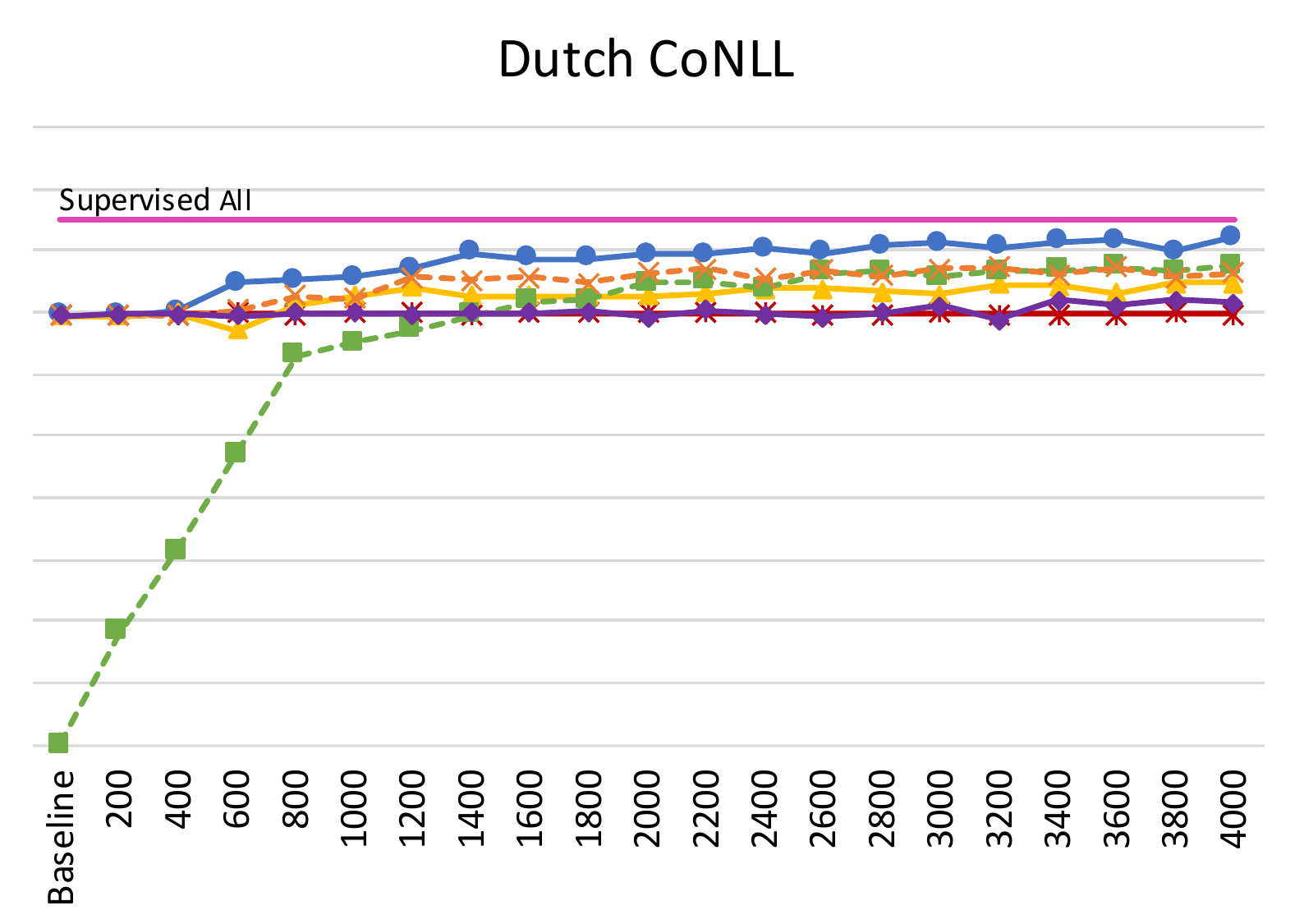}}
~
%\hspace{8pt}%
\subfigure{%
\label{ex3-jb}%
\includegraphics[width=0.5\textwidth]{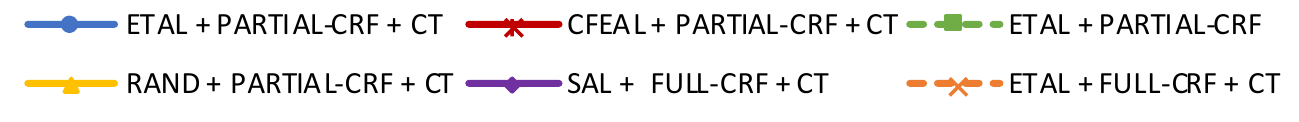}}%
~

  \caption{Comparison of the NER performance trained with the FineTune scheme, across six datasets. Solid lines compare the different active learning strategies. Dashed lines show the ablation experiments. The x-axis denotes the total number of tokens annotated and the y-axis denotes the F1 score.}%
\label{fig:ex3}%
\end{figure*}

Figure \ref{fig:ex3} summarizes the results for all datasets across the different experimental settings. Each data-point on the x-axis corresponds to the NER performance after annotating 200 additional tokens. \textsc{\small{CT}} denotes using cross-lingual transferred data to train the initial NER model for both kick-starting the active learning process and also for fine-tuning the NER model on the newly-acquired data. 
\textsc{\small{Partial-CRF/Full-CRF}} denote the type of CRF decoder used in the NER model. Throughout this paper, we report results averaged across all active learning runs unless otherwise noted. Individual scores are reported in the Appendix \S \ref{moreresults}.

As can be seen in the figure, our proposed recipe \textsc{\small{ETAL+Partial-CRF+CT}} outperforms the previous active learning baselines for all the datasets. Holding the other two components of \textsc{\small{CT}} and \textsc{\small{Partial-CRF}} constant, we conduct experiments to compare the different active learning strategies, which are denoted by the solid lines in Figure \ref{fig:ex3}. We see that \textsc{\small{ETAL}} outperforms the other strategies by a significant margin for both the CoNLL datasets: German (+6.1 F1), Spanish (+5.3 F1), Dutch (+6.3 F1) and the LDC datasets: Hindi (+9.3 F1), Indonesian (+9.0 F1), Spanish (+7.5 F1), at the end of all runs. Furthermore, even with just one-tenth annotated tokens, the proposed recipe is only (avg.) -5.2 F1 behind the model trained using all labeled data, denoted by \textsc{\small{Supervised All}}. Although \textsc{\small{CFEAL}} also selects informative spans, \textsc{\small{ETAL}} outperforms it because \textsc{\small{ETAL}} is optimized to select likely entities, causing more entities to be annotated for Hindi (+43), Indonesian (+207), Spanish-CoNLL (+1579), German (+906), Dutch (+836), except for Spanish-LDC (-184). Despite fully labeled data being adding in \textsc{\small{SAL}}, \textsc{\small{ETAL}} outperforms it because \textsc{\small{SAL}} selects longer sentences with fewer entities: Hindi (-934), Indonesian (-1290), Spanish-LDC (-527), Spanish-CoNLL (-2395), German (-2086), Dutch (-2213). 

From Figure \ref{fig:ex3} we see that \textsc{\small{ETAL}} performs better than the baselines across multiple runs. To verify that this is not an artifact of randomness in the test data, we use a paired bootstrap resampling method, as illustrated in \citet{koehn2004statistical}, to compare \textsc{\small{SAL}}, \textsc{\small{CFEAL}}, \textsc{\small{RAND}} with \textsc{\small{ETAL}}. For each system, we compute the F1 score on randomly sampled 50\% of the data and perform 10k bootstrapping steps at three active learning runs. From Table \ref{tab:ttest} we see that the baselines are significantly worse than \textsc{\small{ETAL}} at 600 and 1200 annotated tokens.
\begin{table}
\small
\begin{center}\resizebox{\columnwidth}{!} {
 \begin{tabular}{c|c|c|c|c|c}
  \small
  \textbf{Dataset} & \textbf{Tokens} & \textbf{ETAL} & \textbf{SAL} & \textbf{RAND} & \textbf{CFEAL}  \\
   \toprule
        Hindi & 200 &  54.8 $\pm$ 2.6 &  49.6 $\pm$ 2.8 &  50.2 $\pm$ 0.4 &  50.4 $\pm$ 2.8 \\
            LDC      & 600 &  64.7 $\pm$ 2.6 &  51.5 $\pm$ 2.9 &  56.1 $\pm$ 2.6 &  54.3 $\pm$ 2.5 \\
                  & 1200 &  69.9 $\pm$ 2.5 &  56.6 $\pm$ 2.7 &  56.8 $\pm$ 2.6 &  64.4 $\pm$ 2.7 \\
  \midrule
        Indonesian &200 &  \textbf{47.4$\pm$ 2.6} &  \textbf{47.9 $\pm$ 2.4} &  \textbf{46.8 $\pm$ 2.3} &  \textbf{48.5 $\pm$ 2.3} \\
           LDC       & 600 &  54.5 $\pm$ 2.4 &  44.5 $\pm$ 2.2 &  47.2 $\pm$ 2.2 &  46.0 $\pm$ 2.3 \\
                  & 1200 &  60.5 $\pm$ 2.3  &  44.7 $\pm$ 2.3 &  51.9 $\pm$ 2.3 &  49.5 $\pm$ 2.3 \\
  \midrule
        Spanish & 200 &  66.3 $\pm$ 3.8 &  62.0 $\pm$ 3.6 &  61.2 $\pm$ 1.2 & 62.5 $\pm$ 3.7 \\
           LDC       & 600 &  65.8 $\pm$ 4.1 &  62.5 $\pm$ 3.7 &  61.9 $\pm$ 2.0 &  63.8 $\pm$ 3.7 \\
                  & 1200 &  78.9 $\pm$ 3.5 &  62.3 $\pm$ 3.6 &  64.6 $\pm$ 4.0 &  68.6 $\pm$ 3.9 \\
  \bottomrule
  \end{tabular}
  }
  \caption{Variance analysis for significance testing of different active learning systems using paired bootstrap resampling. $\pm$ denotes the 95\% confidence intervals. Systems which are not statistically significant than the best system \textsc{\small{ETAL}} are in bold. The CoNLL datasets reflect the same observation, as can be seen in Appendix \S \ref{varianceanalysis}.}
  \label{tab:ttest}
  \end{center}
    \vspace{-2.5mm}
\end{table}

\subsubsection{Ablation Study}
In order to study the contribution of \textsc{\small{CT}} and \textsc{\small{partial-CRF}} in improving the NER performance, we conduct the following ablation, denoted by dashed lines in Figure \ref{fig:ex3}.
 
 \paragraph{\textsc{\small{CT}}:}  We observe that the transferred data from English provides a good start to the NER model: 69.4 (Dutch), 63.0 (Spanish-LDC), 65.7 (Spanish-CoNLL), 54.7 (German), 45.4 (Indonesian), 45.0 (Hindi) F1. As expected, cross-lingual transfer helps more for the languages closely related to English which are Dutch, German, Spanish. For this ablation, we train a \textsc{\small{ETAL+Partial-CRF}} where no transferred data is used. Therefore, to create the seed data, we randomly annotate 200 tokens in the target language and thereafter use \textsc{\small{ETAL}}. We observe that as more in-domain data is acquired, the un-transferred  setting soon approaches the transferred setting \textsc{\small{ETAL+Partial-CRF+CT}} suggesting that an efficient annotation strategy can help close the gap between these two systems with as few as $\sim$1000 tokens (avg.). 

\paragraph{\textsc{\small{Partial-CRF}}:}

We study the effect of using the original CRF (\textsc{\small{Full-CRF}}) instead of the \textsc{\small{Partial-CRF}} for training with partially labeled data. Since the former requires fully labeled sequences, the un-annotated tokens in a sequence are labeled with the model predictions. We see from Figure \ref{fig:ex3} that the \textsc{\small{ETAL+Full-CRF+CT}} performs  worse (avg. -4.1 F1) than \textsc{\small{ETAL+Partial-CRF+CT}}. This is because the \textsc{\small{Full-crf}} significantly hurts the recall, as much as by an average of -11.0 points for Hindi, -1.4 for Indonesian, -7.4 for Spanish-LDC, -3.3 for German, -3.7 for Dutch, -4.8 for Spanish CoNLL.

\subsubsection{Comparison of Training Schemes}
\begin{figure*}%
\centering
\subfigure{%
\label{ex4-aa}%
\includegraphics[width=0.4\textwidth,height=40mm]{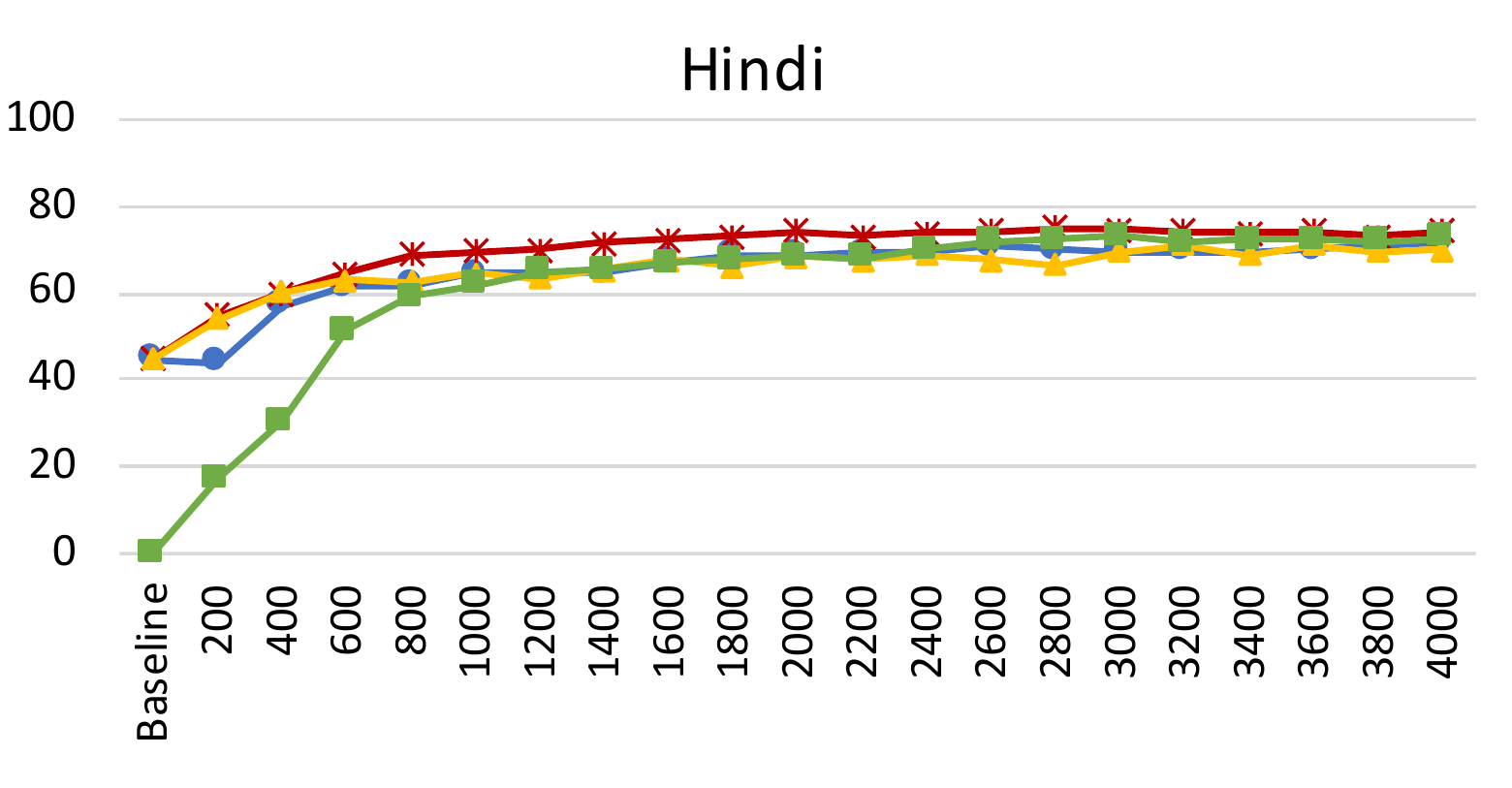}}%\\
\subfigure{%
\label{ex4-aaa}%
\includegraphics[width=0.4\textwidth,height=40mm]{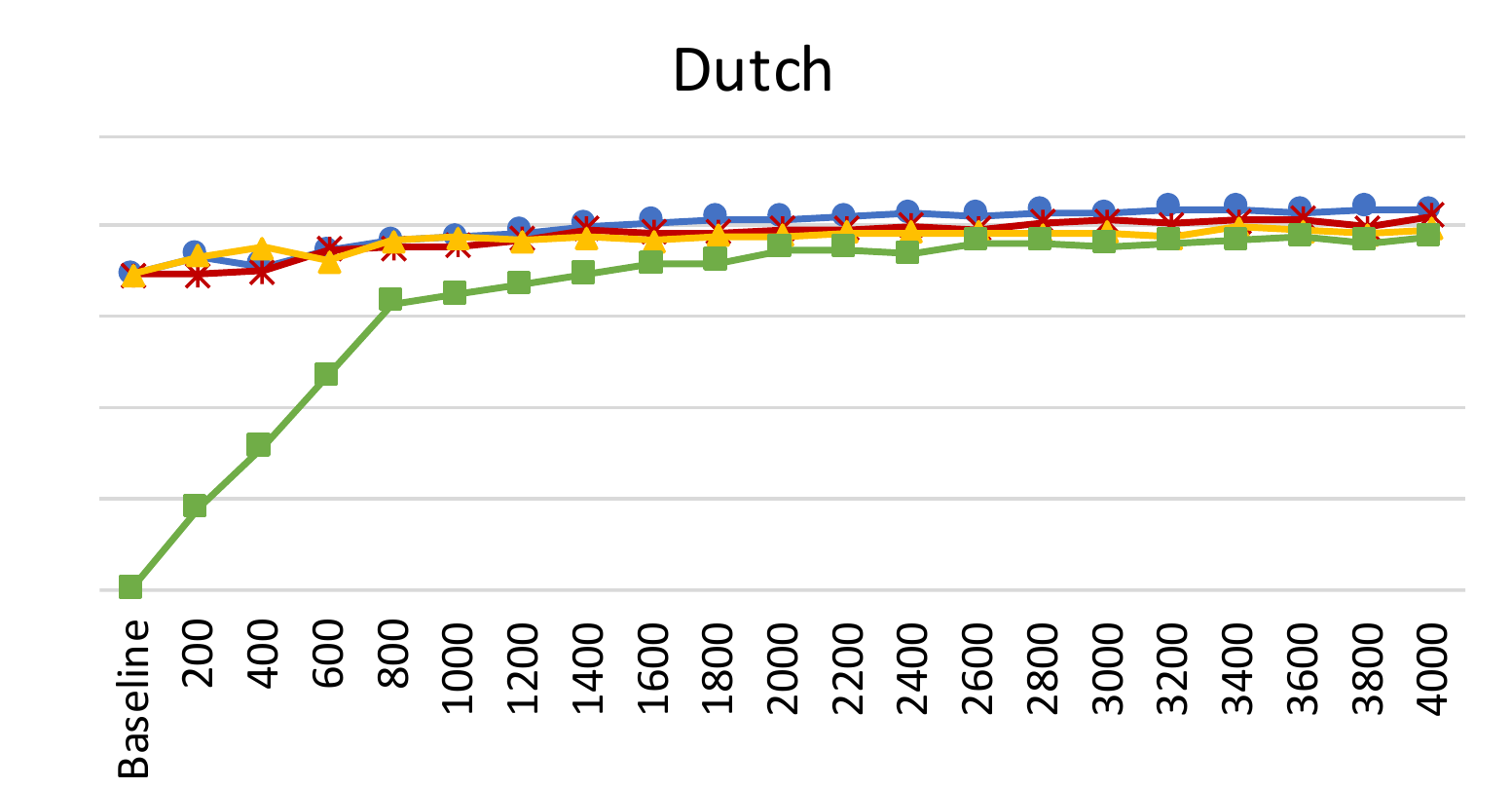}}%\\
\subfigure{%
\label{ex4-aaaa}%
\includegraphics[width=0.2\textwidth,trim={0 0 2cm 0},clip]{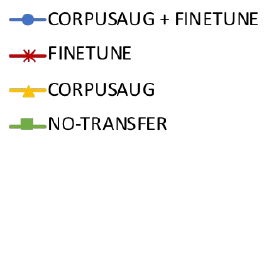}}%\\
~
  \caption{Comparison of the NER performance trained with different schemes for the \textsc{\small{ETAL}} strategy. The x-axis denotes the total number of tokens annotated and the y-axis denotes the F1 score.}%
\label{fig:eg4}%
\end{figure*}
We experiment with different NER training regimes (described in \S \ref{ts}) for \textsc{\small{ETAL}}. We observe that generally fine-tuning not only speeds up the training but also gives better performance than \textsc{\small{CorpusAug}}. For brevity of space, we compare results for two languages in Figure \ref{fig:eg4}:\footnote{The results for other datasets can be found in Appendix \S \ref{trainingscheme}} Dutch, a relative of English, and Hindi, a distant language. We see that \textsc{\small{FineTune}} performs better for Hindi whereas \textsc{\small{CorpusAug+FineTune}} performs better for Dutch. This is because Dutch is closely related to English and benefits the most from the transferred data being explicitly augmented. Whereas for Hindi, which is typologically distant from English, the transferred data is noisy and thus the model doesn't gain much from the transferred data. \citet{xie2018neural} make a similar observation in their experiments with German.

\subsection{Human Annotation Experiments}
\label{human}

\begin{figure*}[h]
\centering
\subfigure[Selected spans using ETAL strategy are highlighted for the human annotator to annotate.]{%
\label{part-1}%
\includegraphics[width=\textwidth]{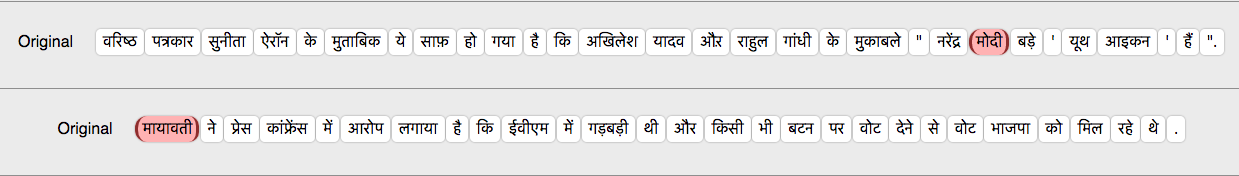}}
~
\subfigure[Human annotator correcting the span boundary and assigning the correct entity type.]{%
\label{part-2}%
\includegraphics[width=\textwidth]{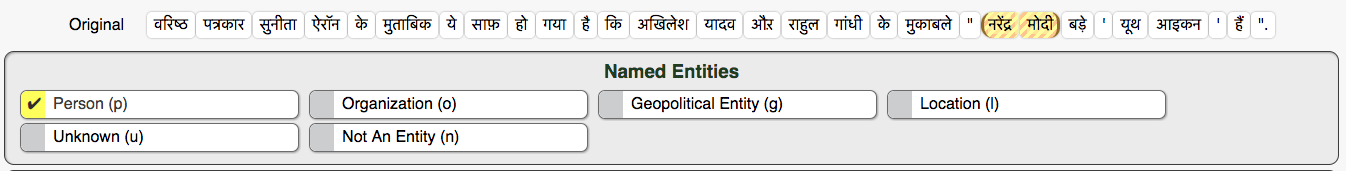}}
~
\subfigure[Human annotator assigning the correct entity type only since selected span boundary is correct.]{%
\label{part-3}%
\includegraphics[width=\textwidth]{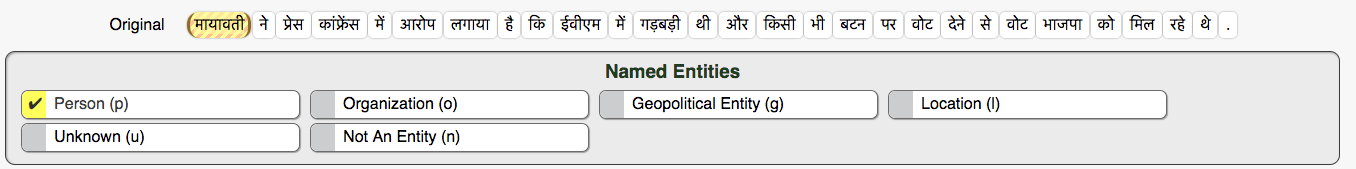}}
~
\subfigure[Partially-annotated sequences after being annotated by the human annotator.]{%
\label{part-4}%
\includegraphics[width=\textwidth]{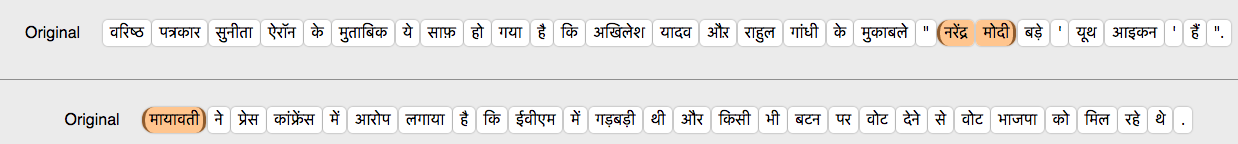}}
~
\caption{Example of the human annotation process for Hindi.}%
\label{fig:interface}%
\end{figure*}

\paragraph{Setup:}  We conduct human annotation experiments for Hindi, Indonesian and Spanish to understand whether \textsc{\small{ETAL}} helps reduce the annotation effort and improve annotation quality in practical settings. We compare \textsc{\small{ETAL}} with the full sequence strategy (\textsc{\small{SAL}}). We use six native speakers, two for each language, with different levels of familiarity with the NER task. Each annotator was provided with practice sessions to gain familiarity with the annotation guidelines and the user interface.  
The annotators annotated for 20 mins time for each strategy. For \textsc{\small{ETAL}}, the annotator was required to annotate single spans i.e each sequence contained one span of tokens.
This involved assigning the correct label and adjusting the span boundary if required. For \textsc{\small{SAL}}, the annotator was required to annotate all possible entities in the sequence. We randomized the order in which the  annotators had to annotate using ETAL and SAL strategy. Figure \ref{fig:interface} illustrates the human annotation process for the ETAL strategy in the annotation interface. \footnote{The code for the annotation interface can be found at \url{https://gitlab.com/cmu_ariel/ariel-annotation-interface.}}

\subsubsection{Results}

\begin{figure*}%
\centering
\subfigure[]{%
\includegraphics[width=0.7\textwidth,trim={3cm 20cm 5cm 1.5cm},clip]{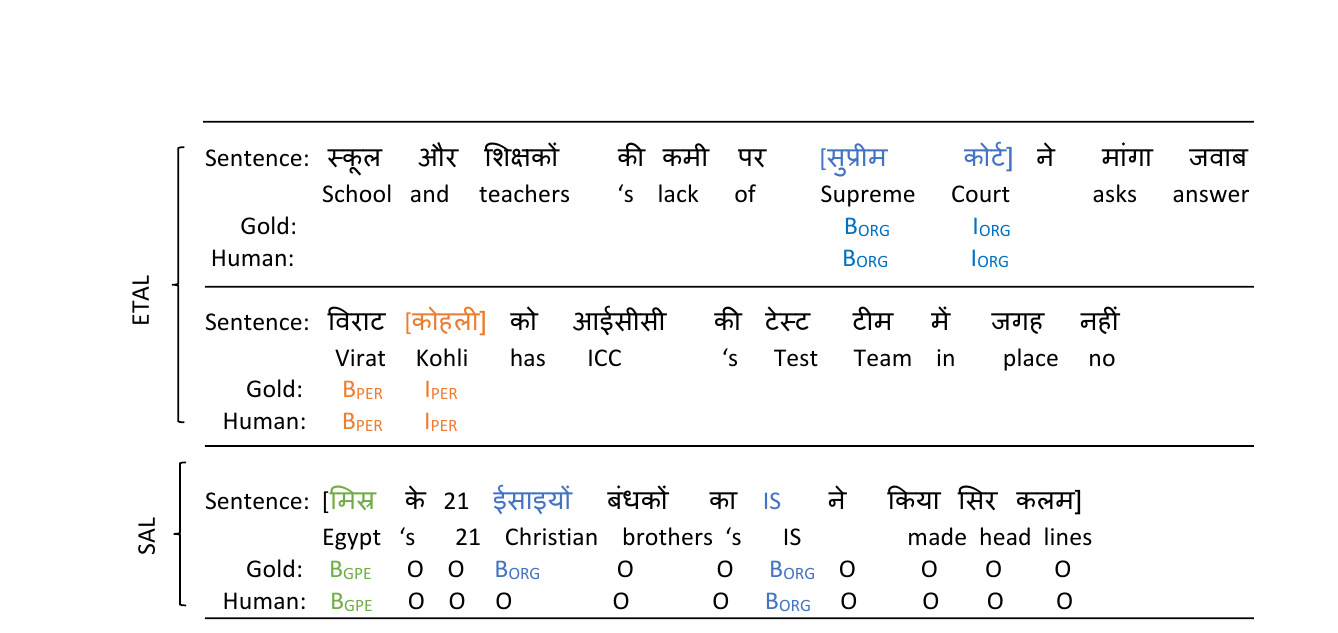} %\llap{
  %\parbox[b]{5cm}{(a)\\\rule{0ex}{2.2in}
  %}} 
  \label{etaleg}
 
  }%\\
~
%\hspace{8pt}%
\subfigure[]{%
\includegraphics[width=0.3\textwidth]{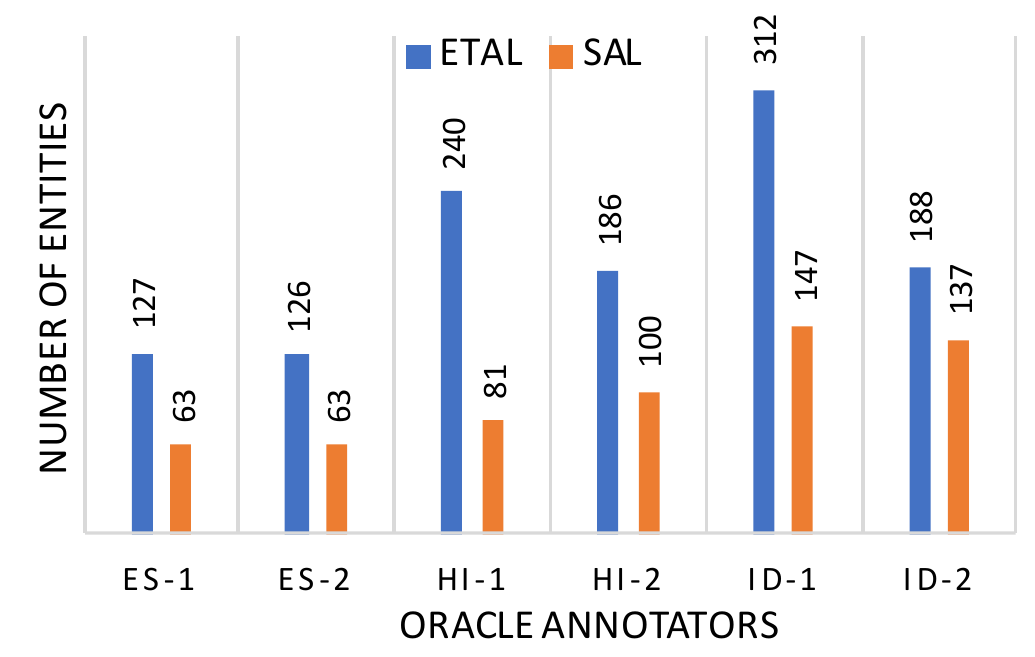} 
%\llap{
 % \parbox[b]{5cm}{(b)\\\rule{0ex}{2.2in}
  %}} 
 \label{entcount} 
 }
~
  \caption{(a) Examples from Hindi human annotation experiments for both \textsc{\small{ETAL}} and \textsc{\small{SAL}}. Square brackets denote the spans (for \textsc{\small{ETAL}}) or the entire sequence (for \textsc{\small{SAL}}) selected by the respective active learning strategy. (b) Comparing the number of entities in the data selected by \textsc{\small{ETAL}} and \textsc{\small{SAL}}, as annotated by oracle. }%
\label{fig:e}%
\end{figure*}

Table \ref{tab:human} records the results of the human annotation experiments. We first compare each annotator's annotation quality with respect to the oracle under both \textsc{\small{ETAL}} and \textsc{\small{SAL}}, denoted by \textit{Annotator Performance}. We see that both  Hindi and Spanish annotators have higher annotation quality using \textsc{\small{ETAL}}. This is because by selecting possible entity spans, \textsc{\small{ETAL}} not only saves effort on searching the entities in a sequence but also allows the annotators to read less overall and concentrate more on the things that they do read, as seen in Figure \ref{etaleg}. However, for \textsc{\small{SAL}} we see that the annotator missed a likely entity because they focused on the other more salient entities in the sequence. For Indonesian, we see an opposite trend due to several inconsistencies in the gold labels. The most common inconsistency being when a common noun is part of an entity. For instance, the gold standard annotates the span \textit{Kabupaten Bogor} as an entity where \textit{Kabupaten} means ``district". Whereas for \textit{ \sout{Kabupaten} Aceh tengah}, the gold standard does not include \textit{Kabupaten}. Similarly, the same span \textit{gunung krakatau} is annotated inconsistently across different mentions where sometimes they exclude the \textit{gunung} (mountain) token.%Moreover, the annotation guideline defines a Reasonable Reader rule which states that ``whether or not any text should be annotated as part of the mention of a name should be decided according to a reasonable, intuitive interpretation of the document". 
Since the annotators referred to these examples during their practice session, their annotations had similar inconsistencies. This issue affects \textsc{\small{ETAL}} more than \textsc{\small{SAL}} because \textsc{\small{ETAL}} selects more entities for annotation. 
\begin{table}
\small
%\begin{center}{
\begin{center}\resizebox{0.5\textwidth}{!}{

   \begin{tabular}{c|c|c|c|c|c}
        %\hline
        & \multicolumn{2}{c|}{\textbf{Annotator}}&\multicolumn{3}{c}{\textbf{Test Performance (\# annotated tokens)}}\\
        
         & \multicolumn{2}{c|}{\textbf{Performance}}&\multicolumn{3}{c}{}\\
      
        \midrule
        &ETAL&SAL&ETAL&SAL&SAL-Full \\ 
        %&&&&&&&(same \# tokens)&\\ 
        \midrule
        
        HI-1 & \textbf{78.8} &	63.7 &  \textbf{50.4} (326)   & 44.2 (326) & 53.3 (1894) \\
        HI-2 & \textbf{82.7} &	72.2 & \textbf{49.1} (234) & 45.9 (234) & 55.6 (2242) \\

        \midrule
        
         ID-1 & 66.1 & \textbf{77.8} & \textbf{50.4} (425) & 45.8 (425) & 51.3 (3232) \\	

         ID-2 & 73.0 & \textbf{79.5} & \textbf{51.2} (251) & 46.5 (251) & 54.0 (2874)     \\
         
         \midrule
         
          ES-1 & \textbf{79.7} &	75.0 & \textbf{63.7} (204) &  62.2 (204) & 64.6 (2134)\\
         ES-2 & \textbf{83.1} &	70.4 & \textbf{63.8} (199) & 62.2 (199) & 62.6 (2134)  \\
         
         \bottomrule
        
    \end{tabular}
  }
  \caption{ Annotator performance measures F1 of each annotator with respect to the oracle. Test Performance measures the NER F1 scores using the annotations as training data. The number in the brackets denote the number of annotated tokens used for training the NER model. ES:Spanish, HI:Hindi, ID: Indonesian.}
  \label{tab:human}
  \end{center}
   \vspace{-7.1mm}
\end{table}

The \textit{Test Performance} compares the performance of the NER models trained with these annotations. The number in the brackets denotes the total number of annotated tokens used for training the NER model. We observe that \textsc{\small{SAL}} has a larger number of annotated tokens than \textsc{\small{ETAL}}. This is because most sequences selected by \textsc{\small{SAL}} did not have any entities. Since ``not-an-entity" is the default label in the annotation interface, no operation is required for annotating these, allowing for more tokens to be annotated per unit times. When we count the number of entities present in the data selected by the two strategies, we see in Figure \ref{entcount} that data selected by \textsc{\small{ETAL}} has a significantly larger number of entities than \textsc{\small{SAL}}, across all the 6 annotation experiments. Therefore, we first compare the NER performance on the same number of annotated tokens. From Table \ref{tab:human} we see that under this setting \textsc{\small{ETAL}} outperforms \textsc{\small{SAL}}, similar to the simulation results. We note that when we consider all the annotated tokens, \textsc{\small{
SAL}} (denoted by \textsc{\small{SAL-Full}}) has slightly better results. However, despite having 6 times fewer annotated tokens, the difference between \textsc{\small{ETAL}} and \textsc{\small{SAL-Full}} is (avg.) 2.1 F1. This suggests that \textsc{\small{ETAL}} can achieve competitive performance with fewer annotations. %On further investigation, we observe that \textsc{\small{ETAL}} has precision-recall imbalance with much higher precision than the recall, causing the F1 to be low \gn{can you give actual numbers for this somewhere? Maybe in the table?}. 
%\gn{While heuristically modifying the biases didn't help fix the unbalanced precision/recall problem, it would still be really nice to fix this if it is indeed the main reason for F-measure being worse... Maybe by up-weighting the cost function of positive tags during training?}

From both the simulation and human experiments, we can conclude that a targeted annotation strategy such as \textsc{\small{ETAL}}  achieves competitive performance with less manual effort while maintaining high annotation quality. Given that \textsc{\small{ETAL}} can help find twice as many entities as \textsc{\small{SAL}}, a potential application of \textsc{\small{ETAL}} can also be for creating a high-quality entity gazetteer under a short time budget.
%which has utility in situations of emergency where prompt response is required in local languages which are often under-resourced.
Since a naive strategy of \textsc{\small{SAL}} allows for more labelled data to be acquired in the same amount of time, in the future we plan to explore mixed-mode annotation where we choose either full sequences or spans for annotation. 

%% file: relatedwork.tex
\section{Related Work}
\label{related}
\paragraph{Cross-Lingual Transfer:} Transferring knowledge from high-resource languages has been extensively used for improving low-resource NER. More common approaches rely on annotation projection methods where annotations in source language are projected to the target language using parallel corpora \cite{zitouni2008mention,ehrmann2011building} or bilingual dictionaries \cite{xie2018neural,mayhew2017cheap}. Cross-lingual word embeddings \cite{bharadwaj2016phonologically,D18-1366}
also provide a way to leverage annotations from related languages.   %More recently, \citet{rahimi2019multilingual} use transfer from several languages by learning from language-specific transfer errors.

\paragraph{Active Learning (AL):} AL has been widely explored for many NLP tasks- NER: \citet{shen2017deep} explore  token-level annotation strategies, \citet{chen2015study} present a study on AL for clinical NER; %POS-tagging: \citet{fang2017model} find most informative word types for annotation,
\citet{baldridge-palmer:2009:EMNLP} evaluate how well AL works with annotator expertise and label suggestions,  \citet{garrette-baldridge:2013:NAACL-HLT} study type and token based strategies for low-resource languages. %dependency-parsing: \citet{li2016active} apply a probabilistic model for annotating syntactic heads.  
\citet{settles2008analysis} present a nice survey on the different AL strategies for sequence labeling tasks, whereas \citet{marcheggiani2014experimental} discuss the strategies for acquiring partially labeled data. \citet{wanvarie2011active,neubig2011pointwise, sperber2014segmentation} show the advantages of training a model on this partially labeled data. All above methods focus on either token or full sequence annotation. 

The most similar work to ours perhaps is that of \citet{fang2017model}, which selects informative word types for low-resource POS tagging. However, their method requires the annotator to annotate single tokens, which is not trivially applicable for multi-word entities in practical settings.
%Moreover, their method does not account for adjusting the span boundary required for multi-word entities. 

%% file: conclusion.tex
\section{Conclusion}
\label{conclusion}
In this paper, we presented a study on how to efficiently bootstrap NER systems for low-resource languages using a combination of cross-lingual transfer learning and active learning. We conducted both simulated and human annotation experiments across different languages and found that: 1) cross-lingual transfer is a powerful tool, constantly beating systems without using transfer;  2) our proposed recipe works the best among known active learning baselines; 3) our proposed active learning strategy saves annotator much effort while ensuring high quality. In future, to account for different levels of annotator expertise, we plan to combine proactive learning \cite{li2017proactive} with our proposed method.

%% file: appendix.tex
\appendix
\section{Appendix}
\label{sec:appendix}
\subsection{Corpus Statistics}
\label{corpustable}
Table \ref{tab:stats} presents the train/dev/test splits used for the NER model training, along with the total number of tokens present in the training data.
\begin{table}[h]
%\small
\begin{center}\resizebox{0.5\textwidth}{!}{
  \begin{tabular}{l|l|l|c}
  \textbf{Source} &\textbf{Dataset}  & \textbf{Train / Dev / Test}& \textbf{Total Tokens} \\
  & &\textbf{\# Sentences} & \textbf{in Train}   \\
   \toprule
   LDC & Hindi &	2570 / 809 / 1592 & 48604 \\
  & Indonesian   & 3181 / 1001 / 1991& 55270\\
   & Spanish  &1398 / 465 / 928& 31799 \\
  \midrule
  CoNLL&Dutch &13274 / 2307 / 4227& 200059 \\
   &German & 12067 / 2849 / 2984&206846 \\
   &Spanish  &  8357 / 1915 / 1517&264715 \\
  \bottomrule
  \end{tabular}
  }
  \caption{ Corpus Statistics.}
  \label{tab:stats}
  \end{center}
\end{table}
\subsection{NER Model Hyperparameters}
\label{nermodel}
 For each language, we train the model with 100d pre-trained GloVe \cite{pennington2014glove} word embeddings trained on Wikipedia and the monolingual text extracted from the train set. We use hidden size of 200 for each direction of the LSTM and a dropout of 0.5. SGD is used as the optimizer with a learning rate of 0.015. During fine-tuning, the NER model is first trained on the transferred data with the above settings. For the first active learning run, the model is fine-tuned on the target language with a lower learning rate of 1e-5 and for each subsequent run, this rate is increased to 0.015.

\subsection{Training Schemes}
\label{trainingscheme}
\begin{figure*}%
\centering
\subfigure{%
\label{ex4-1}%
\includegraphics[width=0.33\textwidth]{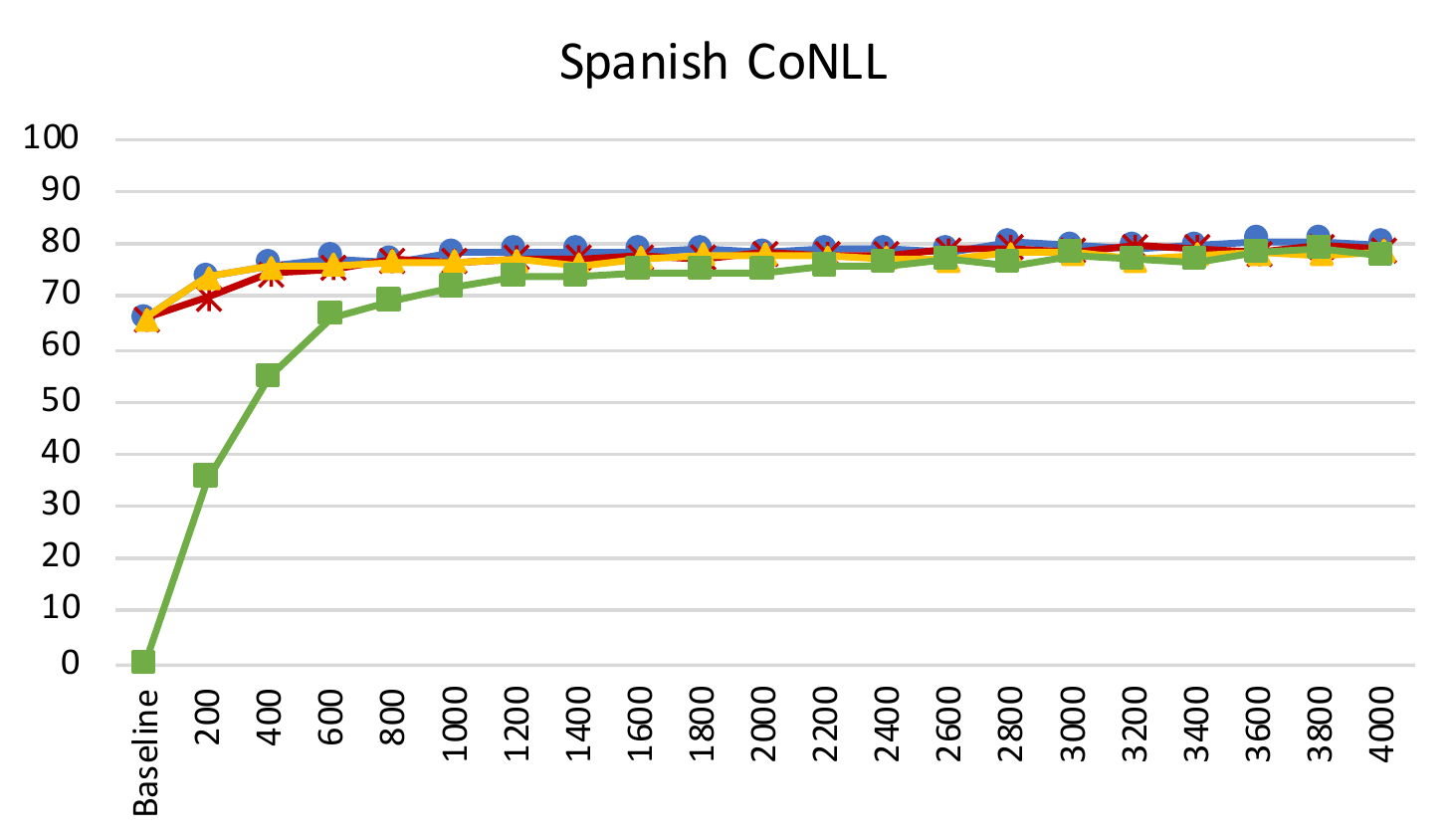}}%\\
~
\subfigure{%
\label{ex4-3}%
\includegraphics[width=0.33\textwidth]{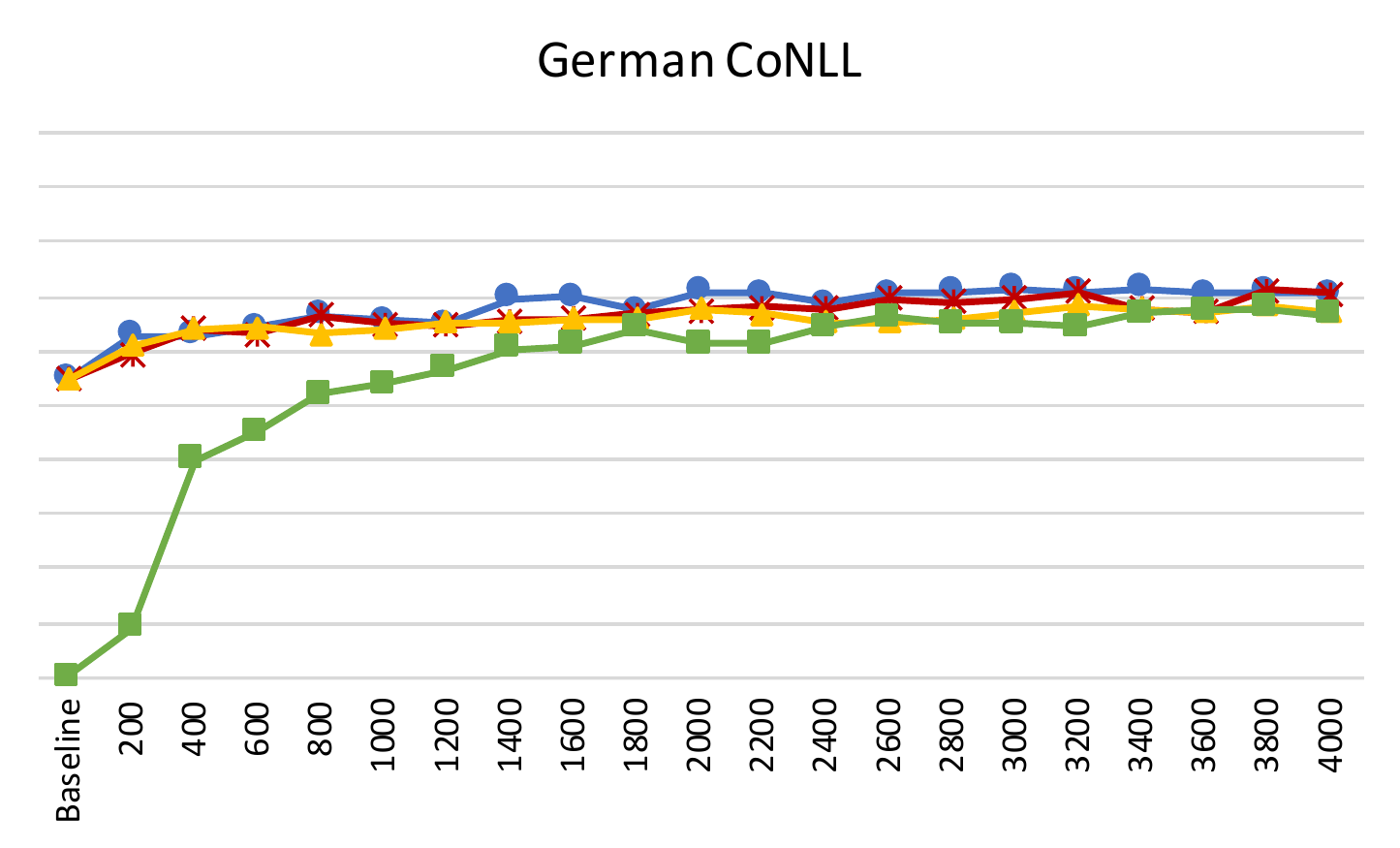}}%\\
~
\subfigure{%
\label{ex4-4}%
\includegraphics[width=0.33\textwidth]{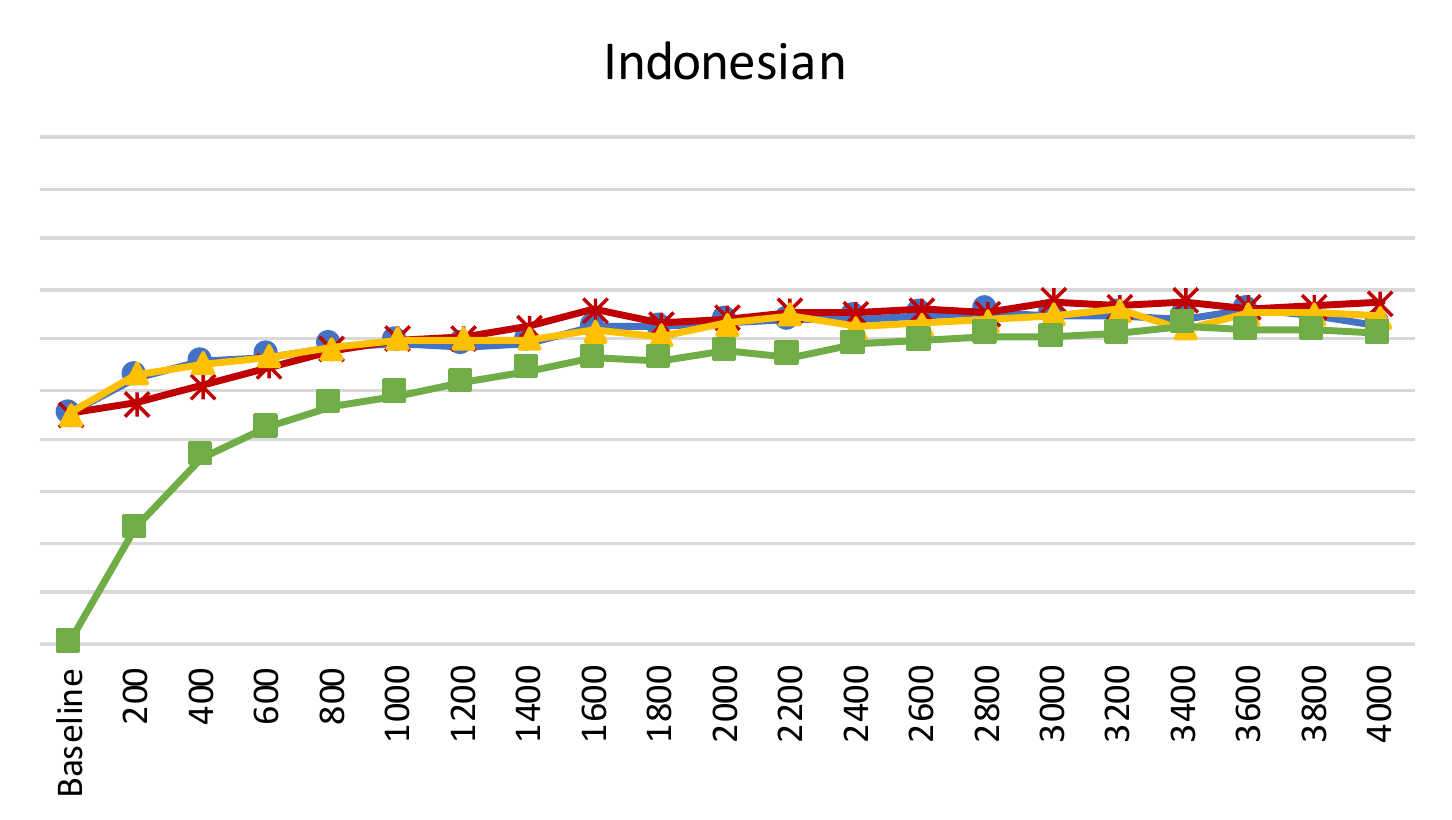}}
~
\subfigure{%
\label{ex4-5}%
\includegraphics[width=0.5\textwidth,trim={0 1.8cm 0 0},clip]{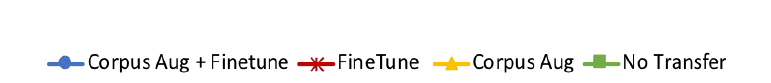}}
~
\caption{Comparison of the NER performance trained with different schemes for the ETAL strategy.}%
\label{fig:ex4}%
\end{figure*}
The results for comparing the different training schemes for Spanish CoNLL, German CoNLL and Indonesian can be seen in Figure \ref{fig:ex4}.

\subsection{Variance Analysis}
\label{varianceanalysis}
Figure \ref{tab:ttestldc} shows the 95\% confidence intervals of the NER models comparing the different active learning strategies for the CoNLL datasets using the bootstrap re-sampling method.

\begin{table}[h]
\small
\begin{center}
\resizebox{\columnwidth}{!}{%
  \begin{tabular}{c|r|c|c|c|c}
  \small
  \textbf{Dataset} & \textbf{Tokens} & \textbf{ETAL} & \textbf{SAL} & \textbf{RAND} & \textbf{CFEAL}  \\
   \toprule

        Dutch & 200 &  \textbf{69.4 $\pm$ 1.6} &  \textbf{69.6                    $\pm$ 1.6} &  \textbf{69.4 $\pm$ 1.6} &                   \textbf{69.4 $\pm$ 1.6} \\
        CoNLL     & 600 &  74.8 $\pm$ 1.6 &  69.4 $\pm$ 1.6 &                        67.2 $\pm$ 2.1 & 66.3 $\pm$ 1.8\\
                  & 1200 & 77.0 $\pm$ 1.5 &  69.6 $\pm$ 1.7 &              74.0 $\pm$ 0.0&  68.7 $\pm$ 1.8\\
   \midrule
        German & 200 &  \textbf{59.3 $\pm$ 1.7} &  \textbf{57.4                $\pm$ 1.9} &  55.2 $\pm$ 2.1 &                  54.7 $\pm$ 2.1 \\
        CoNLL    & 600 &  62.9 $\pm$ 1.7 &  58.7 $\pm$ 1.8 &  58.1                 $\pm$ 2.0 &  57.2 $\pm$ 1.8 \\
                        & 1200 &  64.7 $\pm$ 1.7 &  58.7 $\pm$ 1.8 &  60.7 $\pm$ 1.8 &  60.1 $\pm$ 1.7 \\
   \midrule
        Spanish & 200 &  \textbf{69.7 $\pm$ 1.7} &  65.8 $\pm$ 1.8 &  \textbf{69.5 $\pm$ 1.6}    &  65.3 $\pm$ 1.7 \\
        CoNLL   & 600 &  75.3 $\pm$ 1.8 &  66.3 $\pm$ 1.8 &  73.3 $\pm$ 1.8 &  67.8 $\pm$ 1.7 \\
                  & 1200 &  77.1 $\pm$ 1.7 &  65.7 $\pm$ 1.8 &  73.2 $\pm$ 1.8 &  70.2 $\pm$ 1.7 \\
  \bottomrule
  \end{tabular}
  }
  \caption{Variance analysis for significance testing of different active learning systems using paired bootstrap resampling. $\pm$ denotes the 95\% confidence intervals. Systems which are not statistically significant than the best system \textsc{\small{ETAL}} are highlighted in bold.}
  \label{tab:ttestldc}
  \end{center}
  \vspace{-2.5mm}
\end{table}

\subsection{Comprehensive Results}
\label{moreresults}
Table \ref{tab:hinent}, \ref{tab:indent}, \ref{tab:spaldcent}, \ref{tab:spaldcent}, \ref{tab:deuconllent}, \ref{tab:nedconllent} compares the number of entities present in the data selected by \textsc{\small{ETAL}}, \textsc{\small{CFEAL}} and \textsc{\small{SAL}} across all the datasets.
\begin{table*}[h]
\begin{center}\resizebox{\textwidth}{!}{
  \begin{tabular}{l|l|l|l|l|l|l|l|l|l|l|l|l|l|l|l|l|l|l|l|l|l|l}
    
         \textbf{Method} & \textbf{1} & \textbf{2} & \textbf{3} & \textbf{4} &\textbf{5}&\textbf{6}&\textbf{7}&\textbf{8}&\textbf{9}&\textbf{10}&\textbf{11}&\textbf{12}&\textbf{13}&\textbf{14}&\textbf{15}&\textbf{16}&\textbf{17}&\textbf{18}&\textbf{19}&\textbf{20}\\
      \toprule
        ETAL + PARTIAL-CRF + CT& 115	&192	&281	&379	&482&580	&675	&769	&854	&934	&994	&1083	&1135	&1158	&1171	&1178	&1178	&1179	&1180	&1180 \\
      
        CFEAL+ PARTIAL-CRF + CT&88&207	&298	&397	&506	&608	&698	&793	&877	&978	&1047	&1078	&1104	&1111	&1113	&1119	&1123	&1131	&1132	&1137\\
        
        SAL+ FULL-CRF + CT	&21&42	&45	&52	&60	&70	
	    &88	&95	&111	&126	&133	&150	&158	&174	&184	&195	&210	&227	&235	&246\\
        \bottomrule
  \end{tabular}
  }
  \caption{Comparing number of entities across \textsc{\small{ETAL}}, \textsc{\small{SAL}} and \textsc{\small{CFEAL}} for the Hindi LDC dataset.}
  \label{tab:hinent}
  \end{center}
\end{table*}
\begin{table*}[h]
\begin{center}\resizebox{\textwidth}{!}{
  \begin{tabular}{l|l|l|l|l|l|l|l|l|l|l|l|l|l|l|l|l|l|l|l|l|l|l}
    
         \textbf{Method} & \textbf{1} & \textbf{2} & \textbf{3} & \textbf{4} &\textbf{5}&\textbf{6}&\textbf{7}&\textbf{8}&\textbf{9}&\textbf{10}&\textbf{11}&\textbf{12}&\textbf{13}&\textbf{14}&\textbf{15}&\textbf{16}&\textbf{17}&\textbf{18}&\textbf{19}&\textbf{20}\\
      \toprule
        ETAL + PARTIAL-CRF + CT& 87& 186	&303	&413	&525	&647	&741	&849	&949	&1056	&1138	&1221	&1303	&1360	&1450	&1484	&1511	&1525	&1535	&1536 \\
      
        CFEAL+ PARTIAL-CRF + CT&86&192	&280	&371	&449	&517	&601	&666	&726	&793	&847	&911	&973	&1021	&1069	&1125	&1186	&1244	&1269	&1329\\
        
        SAL+ FULL-CRF + CT	&7& 16	&28	&39	&46	&50	&63	&79	&90	&106	&132	&143	&158	&161	&168	&187	&209	&225	&231	&246\\
        \bottomrule
  \end{tabular}
  }
  \caption{Comparing number of entities across \textsc{\small{ETAL}}, \textsc{\small{SAL}} and \textsc{\small{CFEAL}} for the Indonesian LDC dataset.}
  \label{tab:indent}
  \end{center}
\end{table*}
\begin{table*}[h]
\begin{center}\resizebox{\textwidth}{!}{
  \begin{tabular}{l|l|l|l|l|l|l|l|l|l|l|l|l}
    
         \textbf{Method} & \textbf{1} & \textbf{2} & \textbf{3} & \textbf{4} &\textbf{5}&\textbf{6}&\textbf{7}&\textbf{8}&\textbf{9}&\textbf{10}&\textbf{11}&\textbf{12}\\
      \toprule
        ETAL + PARTIAL-CRF + CT& 84 & 187	&280	&391	&492	&534	&585	&610	&617	&619	&620	&621\\
      
        CFEAL+ PARTIAL-CRF + CT&79& 238	&408	&530	&628	&709	&777	&794	&800	&801	&804	&805\\
        
        SAL+ FULL-CRF + CT	&5& 10	&15	&18	&20	&25	&30	&46	&55	&66	&80	&94\\
        \bottomrule
  \end{tabular}
  }
  \caption{Comparing number of entities across \textsc{\small{ETAL}}, \textsc{\small{SAL}} and \textsc{\small{CFEAL}} for the Spanish LDC dataset.}
  \label{tab:spaldcent}
  \end{center}
\end{table*}
\begin{table*}[h]
\begin{center}\resizebox{\textwidth}{!}{
  \begin{tabular}{l|l|l|l|l|l|l|l|l|l|l|l|l|l|l|l|l|l|l|l|l|l|l}
    
         \textbf{Method} & \textbf{1} & \textbf{2} & \textbf{3} & \textbf{4} &\textbf{5}&\textbf{6}&\textbf{7}&\textbf{8}&\textbf{9}&\textbf{10}&\textbf{11}&\textbf{12}&\textbf{13}&\textbf{14}&\textbf{15}&\textbf{16}&\textbf{17}&\textbf{18}&\textbf{19}&\textbf{20}\\
      \toprule
        ETAL + PARTIAL-CRF + CT& 152& 298&	427	&562	&693	&823	&950	&1094	&1234	&1381	&1503	&1636	&1753	&1882	&2010	&2130	&2257	&2384	&2522	&2674\\
      
        CFEAL+ PARTIAL-CRF + CT&64& 128	&184	&236	&293	&343	&389	&440	&492	&543	&593	&642	&682	&729	&767	&803	&873	&945	&1021	&1095\\
        
        SAL+ FULL-CRF + CT	&27& 44	&66	&79	&88	&102	&117	&129	&132	&142	&154	&172	&180	&196	&223	&232	&240	&252	&263	&279\\
        \bottomrule
  \end{tabular}
  }
  \caption{Comparing number of entities across \textsc{\small{ETAL}}, \textsc{\small{SAL}} and \textsc{\small{CFEAL}} for the Spanish CoNLL dataset.}
  \label{tab:spaconllent}
  \end{center}
\end{table*}
\begin{table*}[h]
\begin{center}\resizebox{\textwidth}{!}{
  \begin{tabular}{ll|l|l|l|l|l|l|l|l|l|l|l|l|l|l|l|l|l|l|l|l|l|l}
    
         \textbf{Method} & \textbf{1} & \textbf{2} & \textbf{3} & \textbf{4} &\textbf{5}&\textbf{6}&\textbf{7}&\textbf{8}&\textbf{9}&\textbf{10}&\textbf{11}&\textbf{12}&\textbf{13}&\textbf{14}&\textbf{15}&\textbf{16}&\textbf{17}&\textbf{18}&\textbf{19}&\textbf{20}\\
      \toprule
        ETAL + PARTIAL-CRF + CT& 154& 264	&386	&513	&664	&775	&883	&1016	&1153	&1275	&1365	&1490	&1588	&1730	&1827	&1954	&2064	&2121	&2211	&2329\\
      
        CFEAL+ PARTIAL-CRF + CT&80	&158	&217	&285	&365	&424	&490	&566	&640	&704	&772	&847	&941	&1008	&1084	&1146	&1220	&1285	&1358	&1423\\
        
        SAL+ FULL-CRF + CT	&22& 68	&74	&81	&93	&101	&112	&123	&135	&148	&166	&175	&188	&198	&205	&213	&224	&230	&239	&243\\
        \bottomrule
  \end{tabular}
  }
  \caption{Comparing number of entities across \textsc{\small{ETAL}}, \textsc{\small{SAL}} and \textsc{\small{CFEAL}} for the German CoNLL dataset.}
  \label{tab:deuconllent}
  \end{center}
\end{table*}
\begin{table*}[h]
\begin{center}\resizebox{\textwidth}{!}{
  \begin{tabular}{l|l|l|l|l|l|l|l|l|l|l|l|l|l|l|l|l|l|l|l|l|l|l}
    
         \textbf{Method} & \textbf{1} & \textbf{2} & \textbf{3} & \textbf{4} &\textbf{5}&\textbf{6}&\textbf{7}&\textbf{8}&\textbf{9}&\textbf{10}&\textbf{11}&\textbf{12}&\textbf{13}&\textbf{14}&\textbf{15}&\textbf{16}&\textbf{17}&\textbf{18}&\textbf{19}&\textbf{20}\\
      \toprule
        ETAL + PARTIAL-CRF + CT& 166& 311	&448	&584	&730	&862	&1008	&1119	&1227	&1356	&1466	&1592	&1708	&1810	&1931	&2041	&2152	&2256	&2376	&2496\\
      
        CFEAL+ PARTIAL-CRF + CT&89	&172	&253	&342	&420	&494	&581	&672	&767	&855	&942	&1020	&1102	&1181	&1259	&1341	&1416	&1505	&1583	&1660\\
        
        SAL+ FULL-CRF + CT	&27& 48	&69	&83	&96	&107	&141	&151	&160	&163	&171	&188	&204	&226	&237	&252	&262	&275	&282	&283\\
        \bottomrule
  \end{tabular}
  }
  \caption{Comparing number of entities across \textsc{\small{ETAL}}, \textsc{\small{SAL}} and \textsc{\small{CFEAL}} for the Dutch CoNLL dataset.}
  \label{tab:nedconllent}
  \end{center}
\end{table*}

Tables \ref{tab:hinres}, \ref{tab:inres}, \ref{tab:sparesldc}, \ref{tab:indores}, \ref{tab:spares},  \ref{tab:nedres}  show the tabulated results for the NER models trained with  different active learning strategies for Hindi, Indonesian, German, Spanish and Dutch datasets.
\begin{table*}[h]
\begin{center}\resizebox{\textwidth}{!}{
  \begin{tabular}{l|l|l|l|l|l|l|l|l|l|l|l|l|l|l|l|l|l|l|l|l|l|l}
    
        \textbf{Type} & \textbf{Method} & \textbf{0} & \textbf{1} & \textbf{2} & \textbf{3} & \textbf{4} &\textbf{5}&\textbf{6}&\textbf{7}&\textbf{8}&\textbf{9}&\textbf{10}&\textbf{11}&\textbf{12}&\textbf{13}&\textbf{14}&\textbf{15}&\textbf{16}&\textbf{17}&\textbf{18}&\textbf{19}&\textbf{20}\\
        \hline
        Span&ETAL + PARTIAL-CRF + CT&45.0&\textbf{54.8}&\textbf{60.0}&\textbf{64.7}&\textbf{68.6}&\textbf{69.7}&\textbf{70.0}&\textbf{71.6}&\textbf{72.3}&\textbf{73.1}&\textbf{74.0}&\textbf{73.2}&\textbf{73.7}&\textbf{74.2}&\textbf{75.1}&\textbf{74.4}&\textbf{74.2}&\textbf{73.8}&\textbf{74.1}&\textbf{73.1}&\textbf{74.3}\\

        &ETAL + PARTIAL-CRF&0.0&17.5&30.3&51.3&59.0&61.7&64.8&65.2&66.8&67.7&68.5&68.0&70.1&72.0&72.5&73.0&71.4&72.1&72.2&72.0&72.8\\

        &ETAL + FULL-CRF + CT&45.0&54.2&	55.8&	57.8&	60.0&	59.5&	61.7&	62.0&	62.5&	63.5 &	63.7&	64.1&	64.2&	64.3&	64.4&	65.2&	65.1&	64.0&	64.9&	64.9&	64.2\\
        \midrule
        Span &CFEAL + PARTIAL-CRF + CT
        &45.0&47.8&46.7&47.4&47.5&60.0&65.5&66.0&67.3&68.0&68.8&69.2&69.6&69.9&70.6&68.8&70.7&71.4&71.1&71.2&72.1\\
        
         &RAND + PARTIAL-CRF + CT&45.0&50.2&53.2&56.1&57.4&56.1&56.9&58.2&59.5&59.5&58.9&60.3&61.7&60.7&61.9&62.4&62.2&62.8&63.3&64.5&65.2\\

        \midrule
        Sequence&SAL + FULL-CRF + CT&45.0&49.6&51.2&51.6&52.6&54.4&56.6&58.6&58.8&59.1&61.2&62.2&60.2&60.1&60.4&60.6&62.7&62.9&62.9&63.1&64.2 \\
        \bottomrule
  \end{tabular}
  }
  \caption{Comparison of NER performance of different active learning strategies for the Hindi LDC dataset. F1 scores are reported. Each column corresponds to NER performance on 200 additional annotated tokens.}
  \label{tab:hinres}
  \end{center}
\end{table*}
\begin{table*}[h]
\begin{center}\resizebox{\textwidth}{!}{
  \begin{tabular}{l|l|l|l|l|l|l|l|l|l|l|l|l|l|l|l|l|l|l|l|l|l|l}
    
        \textbf{Type} & \textbf{Method} & \textbf{0} & \textbf{1} & \textbf{2} & \textbf{3} & \textbf{4} &\textbf{5}&\textbf{6}&\textbf{7}&\textbf{8}&\textbf{9}&\textbf{10}&\textbf{11}&\textbf{12}&\textbf{13}&\textbf{14}&\textbf{15}&\textbf{16}&\textbf{17}&\textbf{18}&\textbf{19}&\textbf{20}\\
        \hline
        Span&ETAL + PARTIAL-CRF + CT&45.4&47.4&50.8&\textbf{54.5}&\textbf{58.0}&\textbf{60.1}&\textbf{60.5}&\textbf{62.3}&\textbf{65.8}&\textbf{63.0}&\textbf{64.0}&\textbf{65.4}&\textbf{65.2}&\textbf{65.7}&\textbf{65.1}&\textbf{67.6}&\textbf{66.7}&\textbf{67.6}&\textbf{66.4}&\textbf{66.7}&\textbf{67.2}\\
        
        &ETAL + PARTIAL-CRF&0.0&22.8&36.8&42.6&47.0&49.1&51.5&53.8&56.3&55.9&57.6&56.6&59.1&59.6&60.8&60.4&61.2&62.7&61.7&61.9&60.9\\
        
       &ETAL + FULL-CRF + CT&45.4&\textbf{48.4} &	\textbf{52.3}&	52.4&	54.2&	54.6&	55.2&	57.0&	57.0&	58.4&	59.1&	59.1&	59.5&	60.7&	60.7&	61.3&	60.3&	60.3&	61.2&	60.9&	60.4\\
        \midrule
        Span&CFEAL + PARTIAL-CRF + CT&45.4&48.5&47.1&46.0&47.5&49.8&49.5&53.9&55.7&54.1&54.9&57.1&55.5&54.7&57.9&56.2&57.9&59.3&58.2&58.6&60.2\\
        
        &RAND + PARTIAL-CRF + CT&45.4&46.8&48.1&47.2&47.2&51.5&51.9&52.5&52.8&52.4&53.2&53.5&54.6&54.1&56.2&55.2&55.8&56.6&58.4&58.6&56.8\\
        \midrule
        
        Sequence&SAL + FULL-CRF + CT&45.4&47.9&45.7&44.5&45.1&45.4&44.7&45.4&48.8&47.8&49.2&50.6&50.3&51.8&51.0&49.9&52.0&51.8&52.4&50.4&52.7\\

        \bottomrule
  \end{tabular}
  }
  \caption{Comparison of NER performance of different active learning strategies for the Indonesian LDC dataset. F1 scores are reported. Each column corresponds to NER performance on 200 additional annotated tokens.}
  \label{tab:inres}
  \end{center}
\end{table*}
\begin{table*}[h]
\begin{center}\resizebox{\textwidth}{!}{
  \begin{tabular}{l|l|l|l|l|l|l|l|l|l|l|l|l|l|l}
    
        \textbf{Type} & \textbf{Method} & \textbf{0} & \textbf{1} & \textbf{2} & \textbf{3} & \textbf{4} &\textbf{5}&\textbf{6}&\textbf{7}&\textbf{8}&\textbf{9}&\textbf{10}&\textbf{11}&\textbf{12}\\
        \hline
        Span&ETAL + PARTIAL-CRF + CT&63.0&\textbf{66.3}&\textbf{67.9}&65.7&\textbf{69.4}&\textbf{74.1}&\textbf{78.9}&\textbf{77.6}&\textbf{78.2}&\textbf{77.6}&\textbf{78.2}&\textbf{76.1}&\textbf{77.2}\\

        &ETAL + PARTIAL-CRF&0.0&10.9&39.8&58.2&63.3&66.8&70.1&70.3&74.5&72.5&72.3&72.5&71.1\\
        
        &ETAL + FULL-CRF + CT&63.0&62.9&	66.6&	\textbf{67.2}&	68.3 &	68.0 &	70.6&	70.1 &	68.5&	69.4&	69.4&	70.6&	69.7\\
        \midrule
        Span&CFEAL + PARTIAL-CRF + CT&63.0&62.5&63.9&63.8&64.1&68.2&68.7&67.2&69.3&68.7&71.9&70.2&70.3\\
        
        &RAND + PARTIAL-CRF + CT&63.0&61.2&61.5&61.9&65.7&65.2&64.6&69.3&67.0&67.3&69.3&69.7&68.7\\
        \midrule
        Sequence&SAL + CT&63.0&62.0&61.8&62.5&61.9&62.3&62.3&62.1&62.3&62.3&62.5&62.7&62.2\\

        \bottomrule
  \end{tabular}
  }
  \caption{Comparison of NER performance (F1 scores) of different active learning strategies for the Spanish LDC dataset. Each column, except Run 0, corresponds to NER performance on 200 additional annotated tokens.} 
  \label{tab:sparesldc}
  \end{center}
\end{table*}
\begin{table*}
\begin{center}\resizebox{\textwidth}{!}{
  \begin{tabular}{l|l|l|l|l|l|l|l|l|l|l|l|l|l|l|l|l|l|l|l|l|l|l}
    
        \textbf{Type} & \textbf{Method} & \textbf{0} & \textbf{1} & \textbf{2} & \textbf{3} & \textbf{4} &\textbf{5}&\textbf{6}&\textbf{7}&\textbf{8}&\textbf{9}&\textbf{10}&\textbf{11}&\textbf{12}&\textbf{13}&\textbf{14}&\textbf{15}&\textbf{16}&\textbf{17}&\textbf{18}&\textbf{19}&\textbf{20}\\
        \hline
        Span & ETAL + PARTIAL-CRF + CT & 54.7
        & 59.3 &	\textbf{64.1}	& 63.0 &	66.5 &	65.0	& 64.7	& 65.4 &	66.0 & 66.8 &\textbf{67.4}	&\textbf{67.9}&\textbf{67.7} & \textbf{69.5} &	\textbf{69.0} & \textbf{69.6} &	\textbf{70.8} &67.7 & 67.1 &\textbf{71.3} &	\textbf{70.5}\\
        
        &  ETAL + PARTIAL-CRF  &0.0 &	9.0 &	39.8 &	45.1 &	51.9 &	53.9 &	56.5 &	60.3 &	61.0 &	64.0 &61.2 &	61.1 & 64.3 & 66.0 &	64.9 & 65.0 &	64.4 &	66.8 &	67.4 &	67.6 &	66.4\\
        
       &ETAL + FULL-CRF + CT&54.7&\textbf{60.7}&63.6&	\textbf{63.9}&	65.4&	\textbf{66.5}&	\textbf{66.6}&	\textbf{66.4}&	\textbf{67.5}&	\textbf{66.9}&	67.3&	66.9&	67.7&	67.7&	68.5&	69.3&	69.3&	\textbf{69.8}&	\textbf{70.7}&	71.0&	70.2\\

        \midrule
         
         Span& CFEAL + PARTIAL-CRF + CT & 54.7 & 54.7 &	55.4 &	57.2 &	59.0 &	61.3 & 	60.2 &	62.3 &	62.1 &	61.4 &	64.5 &	63.9 & 63.5 &63.9 &	65.4 & 65.0 &	66.2 &	65.1 &	65.8 &	65.4 & 66.9 \\
        
         & RAND + PARTIAL-CRF + CT & 54.7 & 55.2 &	57.0 &	58.1 &	59.8 &	57.7 &	60.7 &	59.5 &	57.4 &	57.7 &	59.5 &	60.5 & 58.1 & 59.5 &	61.0 & 58.5 &	58.8 &	60.2	& 61.6 &	61.8   & 58.7 \\
        
        \midrule
        
        Sequence & SAL + FULL-CRF + CT & 54.7 & 57.4 &	57.9 &	58.8 &	58.5 &	59.1 &	58.7 &	58.8 &	58.8 &	59.5 &	57.9 &	57.0 & 56.6 & 60.4 &	60.2 & 60.5 &	61.2 &	60.2 &	61.8 &	60.9 &	60.8 \\
        \bottomrule
  \end{tabular}
  }
  \caption{Comparison of NER performance of different active learning strategies for the German CoNLL dataset. F1 scores are reported. Each column corresponds to NER performance on 200 annotated tokens.}
  \label{tab:indores}
  \end{center}
\end{table*}
\begin{table*}
\begin{center}\resizebox{\textwidth}{!}{
  \begin{tabular}{l|l|l|l|l|l|l|l|l|l|l|l|l|l|l|l|l|l|l|l|l|l|l}
        \textbf{Type} & \textbf{Method} & \textbf{0} & \textbf{1} & \textbf{2} & \textbf{3} & \textbf{4} &\textbf{5}&\textbf{6}&\textbf{7}&\textbf{8}&\textbf{9}&\textbf{10}&\textbf{11}&\textbf{12}&\textbf{13}&\textbf{14}&\textbf{15}&\textbf{16}&\textbf{17}&\textbf{18}&\textbf{19}&\textbf{20}\\
        \hline
        Span&ETAL + PARTIAL-CRF + CT&65.7&69.8&\textbf{74.4}&\textbf{75.3}&\textbf{77.0}&\textbf{76.5}&\textbf{77.1}&\textbf{77.4}&\textbf{77.7}&\textbf{77.2}&\textbf{78.4}&\textbf{78.0}&\textbf{77.9}&\textbf{79.0}&\textbf{79.3}&\textbf{78.7}&\textbf{79.5}&\textbf{79.1}&\textbf{78.3}&\textbf{79.7}&\textbf{79.0}\\
        
        &ETAL + PARTIAL-CRF&0.0&36.4&54.0&64.5&70.5&72.9&72.8&73.7&74.3&75.8&75.2&74.1&76.0&76.2&75.7&76.0&76.5&76.8&76.9&77.2&77.8\\
        
        %  &ETAL + FULL-CRF + CT&65.7& \textbf{72.0}	&68.8	&71.2	&71.7& 72.2& 68.8 &72.2& 72.2& 72.2& 72.2& 72.2& 72.2& 72.2& 72.2& 72.2& 72.2& 72.2& 72.2& 72.2& 72.2\\
         
         &ETAL + FULL-CRF + CT&65.7	&\textbf{72.0}&	68.8	&71.2	&71.7	&72.2	&72.8	&73.3	&73.4	&72.7	&73.3	&74.7	&74.2	&73.9	&73.6	&74.0	&73.9	&74.1	&74.9	&74.5	&73.7\\
        \midrule
        Span&CFEAL + PARTIAL-CRF + CT&65.7&65.3&66.9&67.8&70.9&71.0&70.2&71.6&71.2&73.2&73.2&73.2&72.5&72.7&72.6&72.9&72.0&73.6&73.6&73.4&73.8\\
        
        &RAND + PARTIAL-CRF + CT&65.7&69.5&69.5&70.6&72.1&73.2&70.0&72.0&73.9&73.9&73.6&73.0&71.3&75.7&73.5&74.3&75.1&73.7&74.4&76.2&74.9\\
        \midrule
        Sequence&SAL + FULL-CRF + CT&65.7&65.8&67.4&68.2&68.4&68.2&67.3&67.6&69.4&69.6&69.2&68.9&69.0&69.8&70.0&70.6&71.5&70.7&73.0&70.7&72.7\\

        \bottomrule
  \end{tabular}
  }
  \caption{Comparison of NER performance of different active learning strategies for the Spanish CoNLL dataset. F1 scores are reported. Each column corresponds to NER performance on 200 annotated tokens.}
  \label{tab:spares}
  \end{center}
\end{table*}
\begin{table*}
\begin{center}\resizebox{\textwidth}{!}{
  \begin{tabular}{l|l|l|l|l|l|l|l|l|l|l|l|l|l|l|l|l|l|l|l|l|l|l}
    
        \textbf{Type} & \textbf{Method} & \textbf{0} & \textbf{1} & \textbf{2} & \textbf{3} & \textbf{4} &\textbf{5}&\textbf{6}&\textbf{7}&\textbf{8}&\textbf{9}&\textbf{10}&\textbf{11}&\textbf{12}&\textbf{13}&\textbf{14}&\textbf{15}&\textbf{16}&\textbf{17}&\textbf{18}&\textbf{19}&\textbf{20}\\
        \hline
        Span&ETAL + PARTIAL-CRF + CT&69.4&69.4&\textbf{70.0}&\textbf{74.8}&\textbf{75.2}&\textbf{75.6}&\textbf{77.0}&\textbf{79.4}&\textbf{78.7}&\textbf{78.7}&\textbf{79.2}&\textbf{79.2}&\textbf{80.1}&\textbf{79.5}&\textbf{80.8}&\textbf{81.2}&\textbf{80.4}&\textbf{81.3}&\textbf{81.7}&\textbf{79.8}&\textbf{82.1}\\
        
        &ETAL + PARTIAL-CRF&0.0&18.1&31.4&47.0&62.9&64.9&67.1&69.3&71.7&72.0&74.7&75.0&73.8&76.3&76.5&75.5&76.5&76.7&77.3&76.5&77.5\\
        
        &ETAL + FULL-CRF + CT&69.4&\textbf{69.6}	&69.3	&70.4	&72.6	&72.1	&75.7	&75.1	&75.7	&74.8	&76.3	&76.9	&75.4	&76.8	&75.8	&77.0	&77.3	&76.1	&77.2	&75.7	&76.3\\
        \midrule
        Span&CFEAL + PARTIAL-CRF + CT&69.4&69.5&69.6&69.8&69.6&69.9&69.8&69.7&69.8&69.8&69.8&69.8&69.9&69.6&69.6&69.8&69.6&69.6&69.7&69.7&69.7\\
        
        &RAND + PARTIAL-CRF + CT&69.4&69.5&69.8&67.2&71.3&72.7&74.0&72.5&72.6&72.7&72.5&73.1&73.9&73.8&73.4&72.8&74.4&74.3&73.1&74.6&74.6\\
        \midrule
        Sequence&SAL + FULL-CRF + CT&69.4&\textbf{69.6}&69.7&69.4&69.9&69.8&69.6&69.8&69.9&70.1&69.1&70.3&69.7&69.1&69.9&71.0&68.6&71.9&71.0&71.8&71.4\\

        \bottomrule
  \end{tabular}
  }
  \caption{Comparison of NER performance of different active learning strategies for the Dutch CoNLL dataset. F1 scores are reported. Each column corresponds to NER performance on 200 annotated tokens.}
  \label{tab:nedres}
  \end{center}
\end{table*}

As mentioned in the ablation study which evaluates the effectiveness of \textsc{\small{partial-crf}} over \textsc{\small{full-crf}}, we find that \textsc{\small{full-crf}} significantly hurts the recall. Table \ref{tab:hinre}, \ref{tab:indore}, \ref{tab:spre}, \ref{tab:deure}, \ref{tab:nedre} documents the results of the recall scores across the two settings for Hindi, Indonesian, Spanish-LDC, Spanish-CoNLL, German and Dutch respectively.
\begin{table*}[h]
\begin{center}\resizebox{\textwidth}{!}{
  \begin{tabular}{l|l|l|l|l|l|l|l|l|l|l|l|l|l|l|l|l|l|l|l|l|l}
    
         \textbf{Method} & \textbf{0} & \textbf{1} & \textbf{2} & \textbf{3} & \textbf{4} &\textbf{5}&\textbf{6}&\textbf{7}&\textbf{8}&\textbf{9}&\textbf{10}&\textbf{11}&\textbf{12}&\textbf{13}&\textbf{14}&\textbf{15}&\textbf{16}&\textbf{17}&\textbf{18}&\textbf{19}&\textbf{20}\\
      \toprule
        ETAL + PARTIAL-CRF + CT&38.6&	\textbf{51.4}	& \textbf{56.8} &\textbf{59.7}	&\textbf{60.4}	&\textbf{61.8}	&\textbf{63.2}	&\textbf{64.1}	&\textbf{65.3}	&\textbf{66.8}	&\textbf{68.7}	&\textbf{65.5}	&\textbf{66.9}  &\textbf{67.6}	&\textbf{69.8}	&\textbf{69.9}	&\textbf{71.1} &\textbf{68.1}	&\textbf{68.5}	&\textbf{68.4}	&\textbf{70.7}\\
      
        ETAL + FULL-CRF + CT&38.6&45.8	
        &46.0	&48.3	&50.6	&51.1	&52.6	&53.0	&54.2	&55.6	&55.9	&56.4	&56.4	&54.6	&54.9	&56.6	&56.2	&55.1	&57.3	&57.5	&55.7\\
        \bottomrule
  \end{tabular}
  }
  \caption{Comparing recall scores for evaluating the effectiveness of \textsc{\small{partial-crf}} over \textsc{\small{fullcrf}}  for the Hindi LDC dataset.}
  \label{tab:hinre}
  \end{center}
\end{table*}
\begin{table*}[h]
\begin{center}\resizebox{\textwidth}{!}{
  \begin{tabular}{l|l|l|l|l|l|l|l|l|l|l|l|l|l|l|l|l|l|l|l|l|l}
    
         \textbf{Method} & \textbf{0} & \textbf{1} & \textbf{2} & \textbf{3} & \textbf{4} &\textbf{5}&\textbf{6}&\textbf{7}&\textbf{8}&\textbf{9}&\textbf{10}&\textbf{11}&\textbf{12}&\textbf{13}&\textbf{14}&\textbf{15}&\textbf{16}&\textbf{17}&\textbf{18}&\textbf{19}&\textbf{20}\\
      \toprule
        ETAL + PARTIAL-CRF + CT&51.0 	&47.1	&48.3	&52.1	&55.0	&\textbf{57.6}	&\textbf{61.3}	&\textbf{59.8}	&\textbf{64.3}	&\textbf{61.1}	&\textbf{63.2}	&\textbf{64.0}	&\textbf{64.2}	&\textbf{64.3}	&62.8	&\textbf{66.6}	&\textbf{64.5}	&\textbf{64.6}	&\textbf{63.0}	&\textbf{65.3}	&\textbf{64.2}\\
      
        ETAL + FULL-CRF + CT&51.0	&\textbf{51.3}	&\textbf{55.3}	&\textbf{54.6}	&\textbf{56.6}	&55.4	&56.2	&58.6	&58.9	&60.5	&60.8	&61.1	&61.2	&60.7	&\textbf{63.0}	&62.5	&60.1	&58.9	&62.5	&62.7	&62.3\\
        \bottomrule
  \end{tabular}
  }
  \caption{Comparing recall scores for evaluating the effectiveness of \textsc{\small{partial-crf}} over \textsc{\small{fullcrf}}  for the Indonesian LDC dataset.}
  \label{tab:indore}
  \end{center}
\end{table*}
\begin{table*}
\begin{center}\resizebox{\textwidth}{!}{
  \begin{tabular}{l|l|l|l|l|l|l|l|l|l|l|l|l|l}
    
         \textbf{Method} & \textbf{0} & \textbf{1} & \textbf{2} & \textbf{3} & \textbf{4} &\textbf{5}&\textbf{6}&\textbf{7}&\textbf{8}&\textbf{9}&\textbf{10}&\textbf{11}&\textbf{12}\\
      \toprule
        ETAL + PARTIAL-CRF + CT&57.4	&\textbf{59.5}	&58.5	&57.5	&\textbf{63.9}	&\textbf{72.2}	&\textbf{75.1}	&\textbf{76.0}	&\textbf{76.0}	&\textbf{75.5}	&\textbf{75.6}	&\textbf{73.7}	&\textbf{74.6}   \\
      
        ETAL + FULL-CRF + CT&57.4	&59.4	&\textbf{61.7}	&\textbf{60.6}	&61.5	&61.9	&63.1	&62.1	&62.1	&61.9	&63.3	&63.0 &61.9\\
        \bottomrule
  \end{tabular}
  }
  \caption{Comparing recall scores for evaluating the effectiveness of \textsc{\small{partial-crf}} over \textsc{\small{fullcrf}}  for the Spanish LDC dataset.}
  \label{tab:spre}
  \end{center}
\end{table*}
% \begin{table*}[h]
% \begin{center}\resizebox{\textwidth}{!}{
%   \begin{tabular}{l|l|l|l|l|l|l|l|l|l|l|l|l|l|l|l|l|l|l|1|1|1}
    
%       \textbf{Method} & \textbf{0} & \textbf{1} & \textbf{2} & \textbf{3} & \textbf{4} &\textbf{5}&\textbf{6}&\textbf{7}&\textbf{8}&\textbf{9}&\textbf{10}&\textbf{11}&\textbf{12}&\textbf{13}&\textbf{14}&\textbf{15}&\textbf{16}&\textbf{17}&\textbf{18}&\textbf{19}&\textbf{20}\\
%       \toprule
%         ETAL + PARTIAL-CRF + CT&63.0	&69.2	&74.8	&74.9		&76.4	&76.3	&77.3	&76.7	&77.4		&76.8	&78.0	&77.9	&77.7	&78.6		&79.0	&78.28	&79.2	&79.3	&78.0	&79.0&	78.6 \\
      
%         ETAL + FULL-CRF + CT & 63.0 & 70.5 & & & & & & & & & & & & & & & & & &\\
%         \bottomrule
%   \end{tabular}
%   }
%   \caption{Comparing recall scores for evaluating the effectiveness of \textsc{\small{partial-crf}} over \textsc{\small{fullcrf}}  for the Spanish CoNLL dataset.}
%   \label{tab:spare}
%   \end{center}
% \end{table*}
\begin{table*}
\begin{center}\resizebox{\textwidth}{!}{
  \begin{tabular}{l|l|l|l|l|l|l|l|l|l|l|l|l|l|l|l|l|l|l|l|l|l}
    
       \textbf{Method} & \textbf{0} & \textbf{1} & \textbf{2} & \textbf{3} & \textbf{4} &\textbf{5}&\textbf{6}&\textbf{7}&\textbf{8}&\textbf{9}&\textbf{10}&\textbf{11}&\textbf{12}&\textbf{13}&\textbf{14}&\textbf{15}&\textbf{16}&\textbf{17}&\textbf{18}&\textbf{19}&\textbf{20}\\
      \toprule
        ETAL + PARTIAL-CRF + CT&45.7	&\textbf{58.3}	&\textbf{61.6}	&\textbf{63.9}	&\textbf{63.2}	&\textbf{66.0}	&\textbf{64.1} &\textbf{64.2}	&\textbf{62.8}	&\textbf{65.3}	
        &\textbf{67.4}	&\textbf{67.8}	& \textbf{68.9}	&\textbf{68.1}	&\textbf{69.4}	&\textbf{67.3}	&\textbf{68.4}	&63.5	&63.1	&\textbf{69.5} &65.5 \\
      
        ETAL + FULL-CRF + CT & 45.7 & 52.2 &56.6	&60.2	&61.3	&61.3	&61.1	&62.6	&61.1	&61.0	&61.2	&62.8	&63.1	&63.4	&63.8	&64.4	&65.6	&\textbf{64.2}	&\textbf{64.8} &67.2	&\textbf{65.5}\\
        \bottomrule
  \end{tabular}
  }
  \caption{Comparing recall scores for evaluating the effectiveness of \textsc{\small{partial-crf}} over \textsc{\small{fullcrf}}  for the German CoNLL dataset.}
  \label{tab:deure}
  \end{center}
\end{table*}

\begin{table*}
\begin{center}\resizebox{\textwidth}{!}{
  \begin{tabular}{l|l|l|l|l|l|l|l|l|l|l|l|l|l|l|l|l|l|l|l|l|l}
    
       \textbf{Method} & \textbf{0} & \textbf{1} & \textbf{2} & \textbf{3} & \textbf{4} &\textbf{5}&\textbf{6}&\textbf{7}&\textbf{8}&\textbf{9}&\textbf{10}&\textbf{11}&\textbf{12}&\textbf{13}&\textbf{14}&\textbf{15}&\textbf{16}&\textbf{17}&\textbf{18}&\textbf{19}&\textbf{20}\\
      \toprule
        ETAL + PARTIAL-CRF + CT&65.8	&66.4	&\textbf{70.6}	&\textbf{73.9}	&\textbf{75.1}	&\textbf{75.6}	&\textbf{76.2}	&\textbf{79.4}	&\textbf{78.7}	&\textbf{78.6}	&\textbf{79.1}	&\textbf{78.7}	&\textbf{79.7}	&\textbf{79.0}	&\textbf{80.3}	&\textbf{80.5}	&\textbf{79.7}	&\textbf{81.1}	&\textbf{81.2}	&\textbf{78.9}	&\textbf{81.7} \\
      
        ETAL + FULL-CRF + CT & 65.8 & \textbf{66.9} &66.1	&68.8	&70.9	&70.8	&75.5	&74.1	&75.4	&73.6  &75.6	&76.5	&74.9	&76.1	&75.3	&76.5	&77.0	&75.1   &76.9	&75.1	&75.5\\
        \bottomrule
  \end{tabular}
  }
  \caption{Comparing recall scores for evaluating the effectiveness of \textsc{\small{partial-crf}} over \textsc{\small{fullcrf}}  for the Dutch CoNLL dataset.}
  \label{tab:nedre}
  \end{center}
\end{table*}

% \bibliography{emnlp-ijcnlp-2019}
% \bibliographystyle{acl_natbib}